\documentclass{article}
\usepackage{graphicx}
\usepackage{amsmath}
\usepackage{amssymb}
\usepackage{hyperref}
\usepackage{placeins}
\usepackage[edges]{forest}
\usepackage{geometry}
\usepackage{tikz}
\usepackage{pdflscape}
\usepackage{float}
\usepackage{booktabs}
\usetikzlibrary{shapes, arrows.meta, positioning}
\usepackage{graphicx}
\usepackage{afterpage}   % For placing content on a full page
\usepackage{adjustbox}   % For transformations like flipping

\title{Graph Neural Networks in Multi-Omics Cancer Research: A Structured Survey}

\author{%
	Payam Zohari\textsuperscript{1}\thanks{Email: \href{mailto:payam.zohari@aut.ac.ir}{payam.zohari@aut.ac.ir}} and
	Mostafa Haghir Chehreghani\textsuperscript{1}\thanks{Corresponding author. Email: \href{mailto:mostafa.chehreghani@aut.ac.ir}{mostafa.chehreghani@aut.ac.ir}}
}

\date{\footnotesize\textsuperscript{\textbf{1}}Department of Computer Engineering,
	Amirkabir University
	of Technology (Tehran Polytechnic)\\
	Tehran, Iran}

\begin{document}
%\begin{frontmatter}

\maketitle

%\author[1]{Payam Zohari}
%\email{payam.zohari@aut.ac.ir}
%\author[1]{Mostafa Haghir Chehreghani}
%\email{mostafa.chehreghani@aut.ac.ir}
%\address[1]{Department of Computer Engineering, Amirkabir University of Technology - Tehran Polytechnic, Tehran, Iran}

\begin{abstract}
The task of data integration for multi-omics data has emerged as a powerful strategy to unravel the complex biological underpinnings of cancer. Recent advancements in graph neural networks (GNNs) offer an effective framework to model heterogeneous and structured omics data, enabling precise representation of molecular interactions and regulatory networks. This systematic review explores several recent studies that leverage GNN-based architectures in multi-omics cancer research. We classify the approaches based on their targeted omics layers, graph neural network structures, and biological tasks such as subtype classification, prognosis prediction, and biomarker discovery. The analysis reveals a growing trend toward hybrid and interpretable models, alongside increasing adoption of attention mechanisms and contrastive learning. Furthermore, we highlight the use of patient-specific graphs and knowledge-driven priors as emerging directions. This survey serves as a comprehensive resource for researchers aiming to design effective GNN-based pipelines for integrative cancer analysis, offering insights into current practices, limitations, and potential future directions.
\end{abstract}

\paragraph{keywords}
Data integration, Graph Neural Networks, Multi-omics, Cancer research, Biological tasks

%\end{frontmatter}

\section{Introduction}
\label{sec:intro}

Cancer continues to be a significant global health challenge, accounting for nearly 10 million deaths in 2020, or nearly one in six deaths worldwide \cite{WHO2020}. The complexity of cancer arises from its heterogeneous nature, involving various genetic, epigenetic, and environmental factors that contribute to tumor development and progression.
Early detection and accurate classification of cancer are crucial for effective treatment and improved patient outcomes.
Accordingly, Several recent studies have demonstrated the efficacy of GNN-based  approaches in cancer detection \cite{wu2019comprehensive, shanthamallu2022review}. For instance, a hypergraph convolutional network framework, HyperTMO, was developed to enhance patient classification by capturing complex
multi-omics interactions \cite{10.1093/bib/btae159}. Additionally, a contrastive learning-based GNN model has been introduced to improve the representation of multi-omics
graphs, enabling more accurate phenotype prediction \cite{Rajadhyaksha2023}. Furthermore, a novel tool designed to generate multi-omics signaling graphs, mosGraphGen, facilitates integrative and interpretable graph AI
model development for cancer research \cite{10.1093/bioadv/vbae151}.

Advancements in high-throughput technologies have enabled the comprehensive profiling of various molecular layers, including genomics, transcriptomics, proteomics, and metabolomics. This holistic approach, known as multi-omics, allows for a more detailed understanding of the molecular mechanisms driving cancer. Integrating multi-omics data facilitates the identification of novel biomarkers and therapeutic targets, thereby enhancing the precision of cancer diagnostics and treatments \cite{hasin2017multi,menyhart2021multi,heo2021integrative}.

Despite its potential, integrating multi-omics data poses significant challenges due to the heterogeneity and high dimensionality of the datasets. Traditional computational methods often struggle to capture the complex interactions between different molecular layers, leading to a loss of critical information \cite{bersanelli2016methods}. Therefore, there is a pressing need for advanced computational frameworks capable of effectively integrating and analyzing multi-omics data \cite{misra2019integrated}.

Graph Neural Networks (GNNs) are powerful neural models designed to capture the dependencies within graph-structured data through message passing between nodes, enabling the effective modeling of complex relationships \cite{zhang2020deep, wu2022graph}. They iteratively update node embeddings based on the features of adjacent nodes, allowing for a similarity-preserving mapping where nodes with similar attributes and structural roles are represented by close points in a low-dimensional vector space. This characteristic makes GNNs particularly suitable for various applications, including modeling physical systems, learning molecular fingerprints, predicting protein interfaces, and classifying diseases, where rich relational information is critical \cite{Zhou2020, Khemani2024}. Recent advancements in GNN architectures, such as Graph Convolutional Networks (GCN) and Graph Attention Networks (GAT), have further enhanced their performance across numerous deep learning tasks \cite{Gholamzadeh2024, Lakzaei2024}. By leveraging both structural and content-based embeddings, GNNs provide robust representations that can be utilized in various machine learning algorithms for prediction and classification tasks.

In recent years, a surge of survey literature has emerged focusing on individual and paired intersections of the fields of multi-omics, graph neural networks (GNNs), and cancer research. For instance, the comprehensive review by Lou et al. \cite{lou2024advances} provides an in-depth examination of recent trends in multi-omics data integration and its biological applications, while Hanahan \cite{hanahan2024cancer} offers a contemporary perspective on the evolving understanding of cancer hallmarks. On the computational side, Zhao et al. \cite{zhao2022gnnsurvey} present a detailed survey of GNN methodologies and their broad applications in bioinformatics and systems biology. Several surveys have also explored the intersection of two of these domains.
Rappoport and Shamir \cite{rappoport2020multiomicscancer} analyze multi-omics data integration strategies in the context of cancer, while another paper \cite{gnncancer} examine the role of GNNs in cancer modeling and diagnostics. Additionally, Zhang et al. \cite{zhang2024gnnmultiomics} provide a review of GNN-based approaches for multi-omics data integration. Despite the availability of these valuable resources, to the best of our knowledge, no existing survey provides a comprehensive overview and categorization of GNN-based methods specifically designed for multi-omics data analysis, focusing on cancer research. Given the rapid advancements in GNN-based methodologies for cancer detection using multi-omics data, a comprehensive survey is warranted to synthesize current knowledge, identify existing challenges, and highlight future research directions. 

This survey aims to provide researchers and practitioners with a detailed overview of the state-of-the-art GNN applications in this domain, facilitating further advancements in cancer diagnostics and therapeutics. Our paper fills this critical gap by presenting the first structured review of multi-omics-driven, GNN-based approaches tailored for cancer studies, offering a unified taxonomy and highlighting emerging trends and future research opportunities across this interdisciplinary landscape.
Accordingly, this review introduces a comprehensive framework for understanding multi-omics-driven graph neural network (GNN) approaches in cancer research, emphasizing their innovative applications and integration strategies. We begin by categorizing the field into upstream and downstream tasks. Upstream tasks include omics temporal prediction and the generation of signaling graphs, which leverage GNNs to model dynamic biological processes and inter-omic relationships. Downstream applications focus on cancer classification, subtyping, and the identification of driver genes, where GNNs facilitate the integration of diverse omics data to uncover complex molecular patterns.

Further, we explore various GNN architectures employed in this domain, such as Graph Convolutional Networks (GCN), Graph Attention Networks (GAT), and Graph Transformer Networks (GTN), each offering unique advantages in handling heterogeneous data. Our analysis categorizes existing studies first by the specific tasks they address and then by the GNN models utilized, providing a structured overview of the field's landscape. For each type of GNN used for each individual taks, we discuss the innovation of the papers under that category in detail, besides the types of omics data used—such as genomics, transcriptomics, proteomics, and epigenomics—integrated in the studies, highlighting the diversity of data sources and their contributions to model performance.

Additionally, we introduce key datasets that have been pivotal in advancing multi-omics cancer research, and discuss the evaluation metrics commonly used to assess model. By synthesizing these elements, this survey not only maps the current state of multi-omics GNN applications in cancer research but also identifies emerging trends and challenges, paving the way for future innovations in precision oncology.

The remainder of this paper is organized as follows. In Section \ref{sec:preliminaries}, we define key terms and present the foundational concepts underlying our study. Section \ref{sec:tasks} examines various tasks in the field, offering detailed explanations for each. In Section \ref{sec:gnn-based}, we analyze different GNN models from the literature, highlighting their distinctive characteristics and functionalities. Section \ref{sec:categorization} categorizes the relevant works and discusses the methodologies employed in these studies in depth. Section \ref{sec:datasets} introduces the notable datasets widely utilized in this domain, while Section \ref{sec:metrics} focuses on the evaluation metrics that are critical for performance assessment. In Section \ref{sec:futurework}, we outline several promising avenues for future research aimed at addressing remaining challenges and further advancing the field. To ensure a comprehensive and unbiased synthesis of the literature, our review employs a systematic methodology—detailed in Section \ref{sec:methodology}—that governs the selection and critical appraisal of the studies considered. Finally, Section \ref{sec:conclusion} concludes the paper.

\section{Preliminaries}
\label{sec:preliminaries}

In this section, we define key terminologies and concepts essential for understanding the multiomics-driven graph neural network (GNN) approaches explored in this survey on cancer research.

\paragraph{Graph Neural Networks (GNNs)}
Graph neural networks  \cite{DBLP:conf/iclr/KipfW17,DBLP:conf/nips/HamiltonYL17,DBLP:journals/natmi/Chehreghani22,DBLP:journals/tjs/ZohrabiSC24} are deep learning models tailored to process graph-structured data, where the graph $\mathcal{G} = (\mathcal{V}, \mathcal{E})$ consists of nodes $\mathcal{V}$ and edges $\mathcal{E}$. GNNs iteratively aggregate neighborhood information via message passing to learn node, edge, and global representations.
They are widely used in various applications such as 
fake news detection \cite{DBLP:journals/air/LakzaeiCB24,DBLP:journals/asc/LakzaeiCB25},
recommendation systems \cite{DBLP:journals/corr/abs-2402-03365,DBLP:journals/corr/abs-2502-15699},
bioinformatics \cite{wu2019comprehensive} and medicine and healthcare \cite{10401168,MENG2023107201}.

\paragraph{Omics}
Omics encompasses large-scale biological studies aimed at analyzing molecular-level data across different biological entities (e.g., genes, transcripts, proteins, metabolites). Each omics domain targets a specific layer of molecular biology and provides complementary insights.

\paragraph{Multi-omics}
Multi-omics refers to the integrative analysis of multiple omics layers (e.g., genomics, transcriptomics, proteomics, metabolomics, epigenomics, etc.) to better understand the complexity of biological systems and disease mechanisms. It helps uncover inter-layer relationships and improves biomarker discovery, disease classification, and drug response prediction \cite{hasin2017multi, bersanelli2016methods, rohart2017mixomics, misra2019integrated, subramanian2020multi}.

\paragraph{Different types of omics}

Here, we define some of the omics types that, based on our study, are more popular in multi-omics research. Later in this paper, we use omics type as one of the criteria for comparing existing methods.
% we will compare the f existing work based on these omics.

\begin{itemize}
    \item \textbf{Genomics:} it focuses on the structure, function, and evolution of genomes. It provides information on DNA mutations, single nucleotide polymorphisms (SNPs), and structural variants that influence disease susceptibility \cite{hasin2017multi, subramanian2020multi}. The following is a sample genomic sequence from the \textit{Paramecium bursaria Chlorella virus 1}, available from the NCBI Nucleotide database\footnote{Source: NCBI Nucleotide Database, Accession \texttt{NC\_000852.5},
    	\url{https://www.ncbi.nlm.nih.gov/nuccore/NC_000852}.}:

    \begin{quote}
    \small
    \begin{verbatim}
>NC_000852.5 Paramecium bursaria Chlorella virus 1, complete genome
GGGAGAACCAGGTGGGATTGACAGTGGTAAATGTGTTGACCACGAGTAAAAACAGGGCCCGGAAGCGGGG
CTATATAGAAGAGCGCAAGAAGAACACATAAGGAGAGTTATTTTGATTGGGCAAATCGCTGGCAAAATTG
GCAAAATTTCTTCAATAGTTTTTGTATCAAATAGCGAAAATATTTTTTTTCTCAAAAAGTTTTTTGACTG
GTTAGCGTAAACTATTTTAGTTTCTCATTTATGAGTTTTATGCGAGTTGGTAATAAATCTCACAAAACTC
TAAGGACAAACTCTGGCAGAAATCCTAAGGAGAAACATTAAGAGTTTCGTGGTCTCTGGTCTCGTATCGA
CAAACGAGGTCTTCGGTTGTATATGTTCTCTGGCAATCACTAAGGGCCGCCCGGTGTCATCTGCCGCATC
ACGAAGTACCCCCGAACCCTCAGGTGGTATGTAAACAGAGTATACAAGCTCAAGGACAATTCGTCGATAC
GAAATACTAAGGACATAATTGTTATCTATTTCTTTTAAGTCCGTATTTCGCTTGTTCATAATTTATTGGT
ATAATTTCAGTATTCGCATACTCTTCGGCAAACTTTCTCAAATCATCTAATGAATATTTTTTAGTATTTC
    \end{verbatim}
    \end{quote}

    \item \textbf{Transcriptomics:} it investigates the transcriptome, the complete set of RNA transcripts. It reveals dynamic gene expression profiles under various biological conditions, critical for understanding regulatory mechanisms \cite{hasin2017multi, rohart2017mixomics}.
    A sample of these omics can be seen in Table \ref{tab:transcriptomics}\footnote{Source: 10x Genomics PBMC 3k single-cell RNA-seq dataset, accessed via \url{scanpy.datasets.pbmc3k()}.}.

    \begin{table}[H]
    \centering
    \caption{Transcriptomics data: expression levels of five highly expressed genes across five single cells from the PBMC 3k dataset.}
    \vspace{0.5em} % optional spacing
    \begin{tabular}{lccccc}
    \toprule
    \textbf{Cell ID} & \textbf{RPL13} & \textbf{RPL10} & \textbf{B2M} & \textbf{TMSB4X} & \textbf{MALAT1} \\
    \midrule
    AAACATACAACCAC-1 & 29.0 & 34.0 & 76.0 & 47.0 & 49.0 \\ 
    AAACATTGAGCTAC-1 & 45.0 & 92.0 & 75.0 & 62.0 & 142.0 \\ 
    AAACATTGATCAGC-1 & 16.0 & 49.0 & 69.0 & 117.0 & 171.0 \\ 
    AAACCGTGCTTCCG-1 & 15.0 & 22.0 & 41.0 & 114.0 & 11.0 \\ 
    AAACCGTGTATGCG-1 & 4.0  & 8.0  & 35.0 & 21.0  & 22.0 \\ 
    \bottomrule
    \end{tabular}
    \label{tab:transcriptomics}
    \end{table}

    \item \textbf{Proteomics:} it studies the full set of expressed proteins, their structure, function, and interactions. Proteomics bridges the gap between gene expression and phenotype \cite{hasin2017multi, misra2019integrated}.
    Table \ref{tab:proteomic} presents a sample of protein abundances along with their associated gene symbols\footnote{Source: protein abundance data are often obtained from mass spectrometry or targeted proteomic assays. \url{https://www.nature.com/articles/s41592-020-0840-4}.}.

    \begin{table}[H]
    \centering
    \caption{Sample protein abundances and corresponding gene symbols. \label{tab:proteomic}}
    \begin{tabular}{|l|c|l|}
    \hline
    \textbf{Protein ID} & \textbf{Abundance} & \textbf{Gene Symbol} \\
    \hline
    P12345 & 100.5 & TP53 \\ \hline
    Q67890 & 230.2 & BRCA1 \\ \hline
    R13579 & 89.7  & EGFR \\ 
    \hline
    \end{tabular}
    \end{table}

    \item \textbf{Epigenomics:} it examines epigenetic modifications like DNA methylation and histone modification, which regulate gene expression without changing the DNA sequence. It plays a key role in development and disease \cite{subramanian2020multi}. 
    Table \ref{tab:epigenomic} presents a sample of DNA methylation levels at CpG sites across different chromosomes. Methylation levels range from 0 (no methylation) to 1 (fully methylated)\footnote{ Source: simulated example based on CpG identifiers from the Illumina HumanMethylation450 BeadChip platform.
    	\url{https://support.illumina.com/array/array_kits/infinium_humanmethylation450_beadchip_kit/downloads.html}.}.

    \begin{table}[H]
    \centering
    \caption{Sample methylation levels at selected CpG sites.}
    \begin{tabular}{|l|c|c|}
    \hline
    \textbf{CpG Site} & \textbf{Methylation Level} & \textbf{Chromosome} \\
    \hline
    cg00000029 & 0.85 & chr16 \\ \hline
    cg00000108 & 0.72 & chr7  \\ \hline
    cg00000109 & 0.95 & chr6  \\
    \hline
    \end{tabular}
    \label{tab:epigenomic}
    \end{table}

\end{itemize}

\paragraph{Cancer}
Cancer comprises a group of diseases involving abnormal cell growth with the potential to invade or spread to other parts of the body (metastasis). It results from the accumulation of genetic mutations, epigenetic alterations, and disruptions in molecular signaling pathways \cite{hasin2017multi}.

\section{Tasks}
\label{sec:tasks}

In this section, we divide the tasks studied in the literature into two main categories: downstream tasks and upstream tasks. We describe each category and provide examples from existing works.

\emph{Downstream tasks} involve the direct application of multi-omics data to predict clinical and biological outcomes, including tasks such as classification, survival analysis, and biomarker discovery. These tasks leverage learned representations to forecast patient subtypes, treatment responses, and genetic drivers of cancer.

\emph{Upstream tasks} are dedicated to enhancing data quality, uncovering latent biological structures, and generating graph representations that improve downstream learning. These tasks encompass gene regulatory network inference, omics translation, and synthetic lethality prediction.

Table~\ref{tab:task_classification} presents an overview of the distribution of papers across these categories.

\begin{table}[h]
\centering
\caption{Categorization of related papers based on the tasks they address.}
\label{tab:task_classification}
\resizebox{\textwidth}{!}{ % Resize to half the text width
\begin{tabular}{|l|p{10cm}|}
\hline
\textbf{Task Type} & \textbf{Papers} \\
\hline
\multicolumn{2}{|c|}{\textbf{Downstream Tasks}} \\
\hline
Classification and subtyping & \cite{Ren2024}, \cite{Alharbi2024}, \cite{Alharbi2024LASSO}, \cite{Zhou2024}, \cite{10.1145/3459930.3469542}, \cite{10.1093/bioinformatics/btad353}, \cite{10148642}, \cite{10822233}, \cite{10.1186/s12859-023-05622-4}, \cite{10.1186/s12864-024-11112-5}, \cite{10.1093/bioinformatics/btae523},
\cite{10.3390/ijms25052788}, \cite{10.1093/bib/btae159},\cite{10782177}, \cite{10803430}, \cite{10560503}, \cite{Rajadhyaksha2023}, \cite{10.1093/bib/bbad391}, \cite{10.1093/nargab/lqad063}, \cite{9669359}, \cite{10.1093/bib/bbad081}, \cite{10385631}, \cite{10.3389/fbinf.2023.1164482}, \cite{Yang2023.12.03.569371}, \cite{10.1093/bioinformatics/btad703}, \cite{10385389}, \cite{10.3389/fgene.2022.884028},\cite{HAN2025} \\ \hline
Survival analysis & \cite{Yan2024}, \cite{Zhu2023}, \cite{ZHANG20231}, \cite{10.1093/bioinformatics/btac088}, \cite{9669797}, \cite{10.3390/diagnostics14192178} \\
\hline
Biomarker discovery & 
\cite{Tan2025AMOGEL}, \cite{10.1093/bib/bbae658}, \cite{10.1093/bib/bbaf108}, \cite{Zhang2024.08.01.606222}, \cite{10.3389/fgene.2024.1513938}, \cite{SU2023107252}, \cite{10.1186/s12859-024-06015-x} \\ 
\hline
Driver gene identification & \cite{Chatzianastasis2023}, \cite{Peng2024}, \cite{10.1038/s42256-021-00325-y}, \cite{10.1093/bib/bbae691}, \cite{Dai2024.07.21.604474}, \cite{10831427}, \cite{10822582}, \cite{10.1093/bioinformatics/btac622}, \cite{10385796}, \cite{Yuan2022.09.16.508281}, \cite{10.1093/bioinformatics/btac478}, \cite{10385593}, \cite{10.1093/gigascience/giae108},  \cite{10.1109/JIOT.2025.3526643} \\
\hline
Drug response prediction & \cite{Pu2022}, \cite{LI2025109511}, \cite{10.1186/s12967-024-05976-0}, \cite{10.1038/s41598-024-83090-3}, \cite{10.1186/s12859-024-05987-0}, \cite{LIU2023213}, \cite{10.1093/bib/bbab457}, 
\cite{10.1093/bib/bbab513}, \cite{10.1093/bib/bbad406}  \\
\hline
Pathway/Module discovery & \cite{10629044}, \cite{10.1371/journal.pbio.3002369}, \cite{Fanfani2021.06.29.450342} \\
\hline
\multicolumn{2}{|c|}{\textbf{Upstream Tasks}} \\
\hline
Gene essentiality & \cite{9336285} \\
Synthetic lethality & \cite{10.1093/bioinformatics/btab271} \\
\hline
Omics translation and prediction & \cite{10.1145/3487553.3524714}, \cite{10.1145/3534678.3539213} \\
\hline
Temporal omics prediction & \cite{9932591} \\
\hline
Generate omics signaling graph & \cite{10.1093/bioadv/vbae151} \\
\hline
GRN inference & \cite{Dong2025.01.24.634773}, \cite{Zaki2022PredictingCT} \\
\hline
\end{tabular}
}
\end{table}

\subsection{Definitions of different tasks}
In this section, we present several common tasks in multi-omics research that have been employed in our investigated papers.
% and we also describe each of these tasks briefly.

\begin{itemize}
\item
{\em Classification and subtyping}: This task involves categorizing cells, patients, and tissues based on their molecular profiles, thereby facilitating the development of personalized treatment strategies.

\item 
{\em Survival analysis}: This task involves predicting patient survival outcomes using genomic, transcriptomic, and proteomic data, thereby aiding in the assessment of prognosis.

\item
{\em Biomarker discovery}: This task focuses on identifying key molecular markers that signal the presence or progression of cancer.

\item
{\em Driver gene identification}: This task focuses on detecting specific genes that drive the initiation and maintenance of cancer development.

\item
\emph{Drug response prediction}: This task involves predicting the efficacy of specific drugs for individual patients by analyzing their molecular profiles.

\item 
\emph{Pathway discovery}: This task focuses on identifying molecular pathways or interaction modules that play a role in cancer progression.

\item 
\emph{Gene essentiality}: This task involves identifying the genes that are critical for cell survival, thereby guiding the development of targeted cancer therapies.

\item 
\emph{Synthetic lethality}: This task focuses on identifying pairs of genes for which the concurrent loss of function results in cell death, thereby aiding in cancer drug discovery.

\item 
\emph{Omics translation and prediction}: This task involves mapping the relationships between different omics layers to predict and recover missing data.

\item 
\emph{Temporal omics prediction}: This task focuses on forecasting changes in molecular profiles over time in response to disease progression or treatment.

\item 
\emph{Generate omics signaling graph}: This task involves constructing interpretable graphs that represent multi-omics signaling interactions.

\item 
\emph{GRN inference}: This task involves reconstructing gene regulatory networks to elucidate how genes interact and regulate one another.
\end{itemize}

\section{Types of used GNNs}
\label{sec:gnn-based}

Various studies in the literature utilize a diverse array of graph neural network (GNN) architectures. Table~\ref{tab:gnn_types} classifies these works based on the specific GNN model employed. Below, we provide a brief overview of each GNN type.

\begin{table}[h]
\centering
\caption{Categorization of existing studies based on the GNN architecture.\label{tab:gnn_types}}
\resizebox{\textwidth}{!}{ % Resize to half the text width
\begin{tabular}{|l|p{10cm}|}
\hline
\textbf{GNN type} & \textbf{Paper(s)} \\
\hline
Graph Convolutional Network (GCN) & \cite{Ren2024}, \cite{Pu2022}, \cite{Peng2024}, \cite{10.1145/3459930.3469542}, \cite{ZHANG20231}, \cite{10.1093/bioinformatics/btad353}, \cite{10.1038/s42256-021-00325-y}, \cite{9669797}, \cite{10148642}, \cite{10.1093/bib/bbae691}, \cite{10.1186/s12859-023-05622-4}, \cite{Dai2024.07.21.604474}, \cite{10.1186/s12864-024-11112-5}, \cite{10.1093/bioinformatics/btae523}, \cite{10782177}, \cite{10803430}, \cite{10.1371/journal.pbio.3002369},
\cite{10.1145/3487553.3524714}, 
\cite{9932591}, \cite{Zaki2022PredictingCT}, \cite{10.1145/3534678.3539213}, \cite{Rajadhyaksha2023}, \cite{10.1093/bib/bbab513}, \cite{10.1093/nargab/lqad063}, \cite{9669359}, \cite{10.1093/bib/bbad081}, \cite{10385631}, \cite{10385796}, \cite{10385389}, \cite{10.3389/fgene.2024.1513938}, \cite{10.3389/fgene.2022.884028}, \cite{SU2023107252}, \cite{10.1093/gigascience/giae108}, \cite{HAN2025}, \cite{10.1109/JIOT.2025.3526643} \\ 
\hline
Graph Attention Network (GAT) & \cite{Alharbi2024LASSO}, \cite{10629044}, \cite{10.3390/ijms25052788}, \cite{10.1186/s12859-024-05987-0}, \cite{10.1093/bioinformatics/btac622}, \cite{10560503}, \cite{Yuan2022.09.16.508281}, \cite{Yang2023.12.03.569371},\cite{10.3390/diagnostics14192178}, \cite{9336285} \\ 
\hline
Graph Transformer & \cite{10385593} \\ 
\hline
GraphSAGE & \cite{LI2025109511}, \cite{Dong2025.01.24.634773}, \cite{10.1093/bib/bbaf108} \\ 
\hline
Graph Isomorphism Network (GIN) & \cite{10.1093/bioinformatics/btac478}, \cite{10.1093/bioinformatics/btad703} \\
\hline
Multi-level/Hierarchical GNN & \cite{Yan2024}, \cite{Chatzianastasis2023}, \cite{10.1093/bioinformatics/btac088} \\
\hline
Geometric GNN (GGNN) & \cite{Zhu2023} \\ 
Hypergraph Neural Network (HGNN) & \cite{10.1093/bib/bbae658}, \cite{10.1093/bib/btae159}, \cite{10.1093/bib/bbad391} \\ 
\hline
General GNN & \cite{10822233}, \cite{10.1186/s12967-024-05976-0}, \cite{Zhang2024.08.01.606222}, \cite{10.1093/bioadv/vbae151}, \cite{10822582}, \cite{10.1093/bib/bbab457}, \cite{10.1093/bib/bbad406}, \cite{10.3389/fbinf.2023.1164482}, \cite{Fanfani2021.06.29.450342}, \cite{10.1093/bioinformatics/btab271},
\cite{10.1186/s12859-024-06015-x} \\ 
\hline
Multi-view GNN & \cite{LIU2023213} \\ 
\hline
Hybrid & \cite{Alharbi2024}, \cite{Zhou2024},
\cite{Tan2025AMOGEL}, \cite{10.1038/s41598-024-83090-3}, \cite{10831427} \\ 
\hline
\end{tabular}
}
\end{table}

%\subsection{Descriptions of the used GNN models}
%In this part of the paper, we describe some of the most popular types of GNN used in the field of multi-omics research for cancer study. Each of the GNN types are discussed shortly.
\begin{itemize}
\item
{\em Graph Convolutional Network (GCN)}: 
    GCNs extend traditional convolutional neural networks to graph-structured data by performing localized spectral convolutions on graphs. They aggregate feature information from immediate neighbors to update node representations, efficiently capturing graph topology and node attributes. Introduced by Kipf and Welling, GCNs have become a fundamental building block in many graph learning tasks such as node classification and link prediction \cite{kipf2017semi}.

    \item
    {\em Graph Attention Network (GAT)}: 
    GATs improve upon GCNs by incorporating an attention mechanism that assigns different weights to neighboring nodes during feature aggregation. This dynamic weighting allows the model to focus on more relevant neighbors, enabling better representation learning especially on heterogeneous and noisy graphs \cite{veličković2018graph}.

    \item
    {\em Graph Transformer}: 
    Graph Transformers adapt the self-attention mechanism from Transformers to graph data, enabling the capture of long-range dependencies and global context across nodes. These models overcome the locality limitations of traditional GNNs and have shown promising results in complex graph tasks \cite{dwivedi2021graph}.

    \item
    {\em GraphSAGE}: 
    GraphSAGE is an inductive framework that generates node embeddings by sampling and aggregating features from a fixed-size neighborhood. This approach scales to large graphs and can generalize to unseen nodes by learning aggregation functions \cite{hamilton2017inductive}.

    \item
    {\em Graph Isomorphism Network (GIN)}: 
    GINs aim to achieve maximum expressiveness among GNNs by mimicking the Weisfeiler-Lehman graph isomorphism test. By using sum aggregators with multi-layer perceptrons, GINs can distinguish different graph structures that simpler GNNs cannot \cite{gin}.

    \item
    {\em Multi-level/Hierarchical GNN}: 
    These models apply hierarchical pooling or graph coarsening to capture information at multiple graph resolutions. By progressively compressing graphs into coarser representations, they can capture global graph structure while preserving important local features \cite{multilevelgnn}.

    \item
    {\em Geometric GNN (GGNN)}: 
    GGNNs incorporate geometric and spatial principles, often employing recurrent neural networks to capture temporal or sequential graph data. They excel at modeling structured data with explicit spatial relations, such as molecules or 3D point clouds \cite{geognn}.

    \item
    {\em Hypergraph Neural Network (HGNN)}: 
    HGNNs generalize GNNs to hypergraphs where hyperedges can connect more than two nodes. This extension allows modeling higher-order relationships and interactions beyond pairwise connections, enabling richer graph representations \cite{feng2019hypergraph}.

    \item
    {\em General GNN}: 
    This category refers to the foundational message-passing neural network framework, where node embeddings are iteratively updated by aggregating and transforming messages from neighbors using customizable functions. It provides a flexible basis for various GNN architectures.

    \item
    {\em Multi-view GNN}: 
    Multi-view GNNs integrate multiple graph perspectives or feature sets into a unified model. By leveraging complementary information from different views, they can improve representation learning and downstream task performance \cite{hehuan2022multi}.

    \item 
    {\em Hybrid Approaches}: Techniques that integrate multiple GNN architectures to address different aspects of a problem synergistically.
    
\end{itemize}

%In following section, we provide a detailed literature review of the selected papers. The discussion follows a hierarchical structure, first categorizing papers based on their primary task (downstream or upstream),  and then classifying them by the subtasks, and finally, considering the type of gnn they leverage. For each subtask and methodology, we also proivde type of omics the papers have used as input. This structured analysis provides a comprehensive understanding of GNN applications in multi-omics-driven cancer research.

\section{A categorization of existing work}
\label{sec:categorization}

In this section, we categorize existing work based on the tasks they address. We begin by distinguishing tasks according to their general nature—namely, upstream and downstream. Within these categories, studies are further subdivided by their specific objectives. For each task, the research is subsequently organized according to the type of graph neural network (GNN) employed. Finally, for papers that utilize the same GNN for a particular task, we describe the associated omics type(s) (see Figure~\ref{fig:taxonomy}).

\afterpage{%
%  \clearpage
  \thispagestyle{empty} % Remove page number
  \noindent
  \adjustbox{angle=90, width=\paperheight, height=\paperwidth, keepaspectratio}{
    \includegraphics{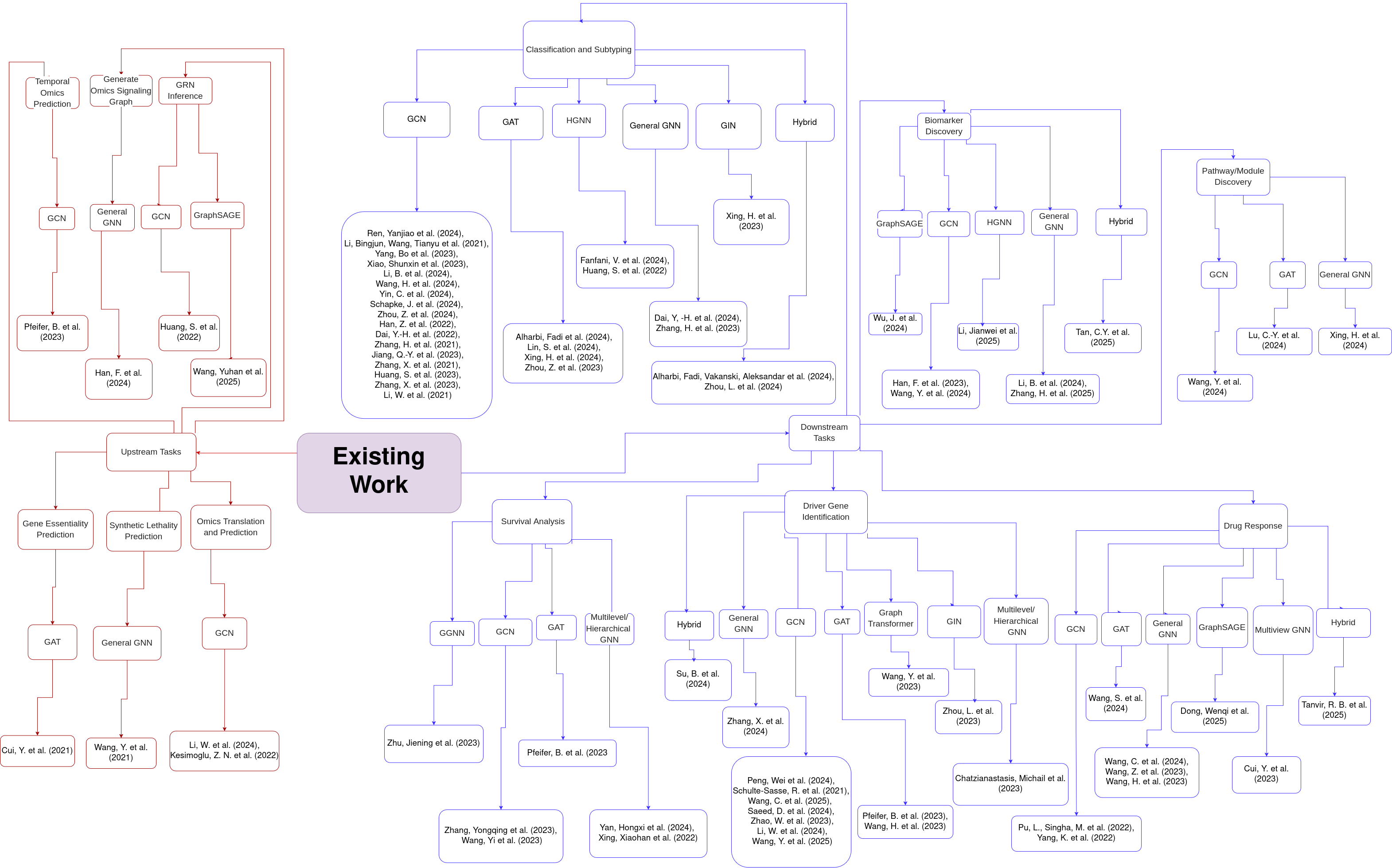}
  }
  \label{fig:taxonomy}
%  \clearpage
}

\subsection{Upstream tasks}
Upstream tasks form the backbone of multi-omics cancer studies, encompassing the initial data transformations and biological relationship modeling that pave the way for downstream phenotypic predictions. These tasks include predicting gene essentiality and synthetic lethality, reconstructing gene regulatory networks, and translating between omics modalities. Within the context of graph neural networks (GNNs), upstream tasks typically involve mapping omics data onto graph structures, thereby enabling the extraction of structured, meaningful information that captures gene dependencies, interactions, and temporal dynamics. This section reviews key upstream tasks and examines the GNN methods employed for each.

\subsubsection{Gene essentiality prediction}
Gene essentiality prediction aims to identify genes that are critical for cellular viability. In cancer research, isolating these essential genes facilitates the discovery of therapeutic targets that, when inhibited, induce selective cancer cell death. Multi-omics features—such as gene expression profiles and protein–protein interaction (PPI) networks—offer rich signals for discerning these dependencies (see Table~\ref{tab:gnn_omics_gene_essentiality}).

\begin{table}[H]
\centering
\caption{GNN types and omics data used for gene essentiality prediction.}
\begin{tabular}{|c|c|cccc|}
\hline
\textbf{Paper} & \textbf{GNN Type} & \multicolumn{4}{c|}{\textbf{Omics Types}} \\
\cline{3-6}
 &  & \textbf{Genomic} & \textbf{Epigenomic} & \textbf{Proteomic} & \textbf{Transcriptomic} \\
\hline
\cite{9336285} & GAT &  & \checkmark &  & \checkmark \\
\hline
\end{tabular}
\label{tab:gnn_omics_gene_essentiality}
\end{table}

\paragraph{GAT}
In \cite{9336285}, the authors introduce EPGAT, a Graph Attention Network (GAT)-based framework for predicting essential genes across multiple organisms, including human datasets pertinent to cancer. The method formulates gene essentiality as a node classification problem on attributed protein–protein interaction (PPI) networks, where each node corresponds to a gene enriched with omics features. By integrating self-attention into the message passing process, EPGAT dynamically assesses the significance of neighboring nodes, enabling adaptive learning of regulatory importance. The framework leverages a suite of multi-omics signals—such as gene expression, orthology, and network topology—to outperform traditional classifiers, particularly under incomplete label conditions. Its primary innovation lies in directly exploiting graph structures without the need for tabular preprocessing, thereby facilitating cross-species generalization.

\subsubsection{Synthetic lethality prediction}
Synthetic lethality (SL) refers to a condition in which the simultaneous loss of both genes in a pair is lethal to a cell, whereas the disruption of either gene individually is tolerable. This concept offers considerable therapeutic potential, especially in cancers harboring specific mutations, as targeting the SL partner gene can selectively eliminate tumor cells (see Table~\ref{tab:gnn_omics_synthetic_lethality}).

\begin{table}[H]
\centering
\caption{GNN types and omics data used for synthetic lethality prediction.}
\begin{tabular}{|c|c|cccc|}
\hline
\textbf{Paper} & \textbf{GNN Type} & \multicolumn{4}{c|}{\textbf{Omics Types}} \\
\cline{3-6}
 &  & \textbf{Genomic} & \textbf{Epigenomic} & \textbf{Proteomic} & \textbf{Transcriptomic} \\
\hline
\cite{10.1093/bioinformatics/btab271} & General GNN &  & \checkmark &  & \checkmark \\
\hline
\end{tabular}
\label{tab:gnn_omics_synthetic_lethality}
\end{table}

\paragraph{General GNN}
In \cite{10.1093/bioinformatics/btab271}, the authors introduce the KG4SL framework, which integrates knowledge graphs with a graph neural network (GNN) architecture to predict synthetic lethality (SL) interactions. The approach constructs gene-specific subgraphs from a large-scale biological knowledge graph that incorporates diverse entity types and relationships—such as genes, diseases, and drugs. By performing message passing over these heterogeneous subgraphs, the model learns biologically contextualized gene embeddings. A subsequent decoder then computes similarity scores between gene pairs to predict SL interactions. A key contribution of KG4SL is its ability to automatically capture intricate gene–gene dependencies across multiple biological domains without manual feature engineering, thereby achieving strong predictive performance in human cancers.

\subsubsection{Omics translation and prediction}
Omics translation tasks focus on predicting one omic modality (e.g., miRNA expression) from another (e.g., methylation or mRNA expression). This approach is vital for compensating for missing modalities and inferring unmeasured signals in practical biomedical applications (see Table~\ref{tab:gnn_omics_translation}).

\begin{table}[H]
\centering
\caption{GNN types and omics data used for omics translation.}
\begin{tabular}{|c|c|cccc|}
\hline
\textbf{Paper} & \textbf{GNN Type} & \multicolumn{4}{c|}{\textbf{Omics Types}} \\
\cline{3-6}
 &  & \textbf{Genomic} & \textbf{Epigenomic} & \textbf{Proteomic} & \textbf{Transcriptomic} \\
\hline
\cite{10.1145/3487553.3524714} & GCN &  & \checkmark &  & \checkmark \\
\hline
\cite{10.1145/3534678.3539213} & GCN &  & \checkmark &  & \checkmark \\
\hline
\end{tabular}
\label{tab:gnn_omics_translation}
\end{table}

\paragraph{GCN}
In \cite{10.1145/3534678.3539213}, the authors propose **scMoGNN**, a Graph Convolutional Network (GCN)-based framework for integrating multimodal single-cell data. In this approach, both cells and features (e.g., genes and chromatin markers) are represented as nodes in a bipartite graph, with edges encoding cell-feature associations. A tailored GCN module propagates structural and biological information throughout the graph, thereby enhancing cell-level representations for tasks such as modality prediction and imputation. By incorporating external biological priors (e.g., gene–gene relationships), scMoGNN enables accurate translation across modalities and significantly outperforms previous integration techniques, particularly in low-annotation regimes.
Similarly, in \cite{10.1145/3487553.3524714}, the authors introduce a multiplex graph neural network framework that constructs multiple intra- and inter-omics similarity graphs. In this model, each layer of the multiplex graph is processed by individual GCNs, with the outputs subsequently combined via an attention mechanism. An attention-based encoder is applied to each graph view, and the resulting node

\subsubsection{Temporal omics prediction}
Temporal omics modeling is instrumental in forecasting the evolution of molecular profiles over time, thereby supporting trajectory inference in contexts such as cancer progression and drug response. By integrating temporal dynamics with graph-based biological priors, these models achieve enhanced robustness and interpretability (see Table~\ref{tab:gnn_omics_temporal}).

\begin{table}[H]
\centering
\caption{GNN types and omics data used for temporal omics prediction.}
\begin{tabular}{|c|c|cccc|}
\hline
\textbf{Paper} & \textbf{GNN Type} & \multicolumn{4}{c|}{\textbf{Omics Types}} \\
\cline{3-6}
 &  & \textbf{Genomic} & \textbf{Epigenomic} & \textbf{Proteomic} & \textbf{Transcriptomic} \\
\hline
\cite{9932591} & GCN &  & \checkmark &  & \checkmark \\
\hline
\end{tabular}
\label{tab:gnn_omics_temporal}
\end{table}

\paragraph{GCN}
\cite{9932591} introduces NeTOIF, a hybrid model that integrates Graph Convolutional Networks (GCN) with Long Short-Term Memory (LSTM) networks to forecast temporal omics values in longitudinal datasets. The GCN component encodes gene and protein interaction topologies, capturing structural dependencies, while the LSTM module models temporal patterns. Together, these components enable robust imputation of missing data points and accurate prediction of future values across modalities such as proteomics and transcriptomics. NeTOIF is distinguished by its dynamic approach to handling time-evolving missing values, achieving improved accuracy over classical regression methods and other deep learning baselines.

\subsubsection{Generate omics signaling graph}
This task involves generating biologically informed signaling or functional graphs directly from raw omics data. Such graphs capture intricate molecular interactions and serve as the structural backbone for various downstream learning tasks in cancer biology (see Table~\ref{tab:gnn_omics_signaling_graph}).

\begin{table}[H]
\centering
\caption{GNN types and omics data used for signaling graph generation.}
\begin{tabular}{|c|c|cccc|}
\hline
\textbf{Paper} & \textbf{GNN Type} & \multicolumn{4}{c|}{\textbf{Omics Types}} \\
\cline{3-6}
 &  & \textbf{Genomic} & \textbf{Epigenomic} & \textbf{Proteomic} & \textbf{Transcriptomic} \\
\hline
\cite{10.1093/bioadv/vbae151} & General GNN & \checkmark & \checkmark & \checkmark & \checkmark \\
\hline
\end{tabular}
\label{tab:gnn_omics_signaling_graph}
\end{table}

\paragraph{General GNN}
In \cite{10.1093/bioadv/vbae151}, the authors propose the mosGraphGen framework, designed to construct interpretable signaling graphs from multi-omics cancer data. This method maps omics profiles onto biologically structured graphs that encode known gene–gene interactions, expression dynamics, and pathway annotations. The resulting graphs are subsequently fed into Graph Neural Networks (GNNs) for downstream tasks, such as classification or biomarker discovery. A key innovation of mosGraphGen is its flexible graph construction pipeline, which produces learning-ready structures while preserving biological interpretability.

\subsubsection{Gene Regulatory Network (GRN) Inference}
GRN inference reconstructs the intricate network of regulatory relationships between genes, providing a comprehensive systems-level perspective on transcriptional regulation and its perturbation in cancer (see Table~\ref{tab:gnn_omics_grn}).

\begin{table}[H]
\centering
\caption{GNN types and omics data used for GRN inference.}
\begin{tabular}{|c|c|cccc|}
\hline
\textbf{Paper} & \textbf{GNN Type} & \multicolumn{4}{c|}{\textbf{Omics Types}} \\
\cline{3-6}
 &  & \textbf{Genomic} & \textbf{Epigenomic} & \textbf{Proteomic} & \textbf{Transcriptomic} \\
\hline
\cite{Zaki2022PredictingCT} & GCN &  & \checkmark &  & \checkmark \\
\hline
\cite{Dong2025.01.24.634773} & GraphSAGE &  & \checkmark &  & \checkmark \\
\hline
\end{tabular}
\label{tab:gnn_omics_grn}
\end{table}

\paragraph{GCN}
In \cite{Zaki2022PredictingCT}, the authors present a GCN-based framework for inferring gene regulatory networks (GRNs) from single-cell RNA-seq data. By modeling gene co-expression as a graph structure, the approach captures both local regulatory relationships (such as transcription factor–gene interactions) and broader, module-level associations. The graph is initially constructed using established gene interaction data, and omics features are subsequently propagated through the GCN to iteratively refine the inferred links. The output is a robust GRN that highlights key transcriptional regulators alongside their downstream targets, offering valuable insights into the underlying regulatory mechanisms in cancer.

\paragraph{GraphSAGE}
In \cite{Dong2025.01.24.634773}, the authors introduce scGraphETM, a scalable framework that combines GraphSAGE with embedded topic modeling to infer gene regulatory networks (GRNs) from single-cell multi-omics data. The method constructs a heterogeneous graph that links transcription factors (TFs), regulatory elements, and genes by integrating chromatin accessibility with gene expression profiles. GraphSAGE enables the efficient aggregation of neighborhood information, while the topic model enhances interpretability by mapping latent topics to distinct biological processes. The result is a cell-type-specific GRN that offers both improved biological relevance and enhanced computational efficiency.

\subsection{Downstream tasks}
Downstream tasks in multi-omics-driven cancer analysis focus on deriving clinically actionable insights from structured graph representations. Following the upstream transformation of omics data into rich graph structures, GNN-based models are employed for a range of predictive tasks, including disease classification, prognosis, biomarker discovery, and treatment response prediction. These applications are crucial for advancing precision medicine by enhancing patient stratification and informing therapy selection. In this section, we explore how various GNN variants have been leveraged across these core downstream applications.

\subsubsection{Classification and subtyping}
Classification and subtyping tasks are pivotal in delineating cancer types or molecular subgroups based on patient omics profiles. Effective subtyping is essential for capturing disease heterogeneity and enhancing treatment outcomes. Graph Neural Networks (GNNs) are particularly adept at this, as they learn graph-aware feature embeddings that reveal subtle dependencies among omics features or patient relationships. By leveraging the intrinsic relational structure of biological data, GNN-based models enable more precise disease classification and facilitate personalized therapeutic strategies (see Table~\ref{tab:classification_subtyping}).

\begin{table}[H]
\centering
\caption{GNN Types for classification and subtyping.}
\begin{tabular}{|c|p{10cm}|}
\hline
\textbf{GNN Type} & \textbf{Paper(s)} \\
\hline
GCN & \cite{Ren2024}, \cite{10.1145/3459930.3469542},                            \cite{10.1093/bioinformatics/btad353}, 
      \cite{10148642}, \cite{10.1186/s12859-023-05622-4}, 
      \cite{10.1186/s12864-024-11112-5},
      \cite{10.1093/bioinformatics/btae523}, 
      \cite{10782177}, \cite{10803430}, 
      \cite{Rajadhyaksha2023}, \cite{10.1093/nargab/lqad063}, 
      \cite{9669359}, \cite{10.1093/bib/bbad081}, 
      \cite{10385631}, \cite{10385389}, 
      \cite{10.3389/fgene.2022.884028}, 
      \cite{HAN2025} \\
\hline
GAT & \cite{Alharbi2024LASSO}, \cite{10.3390/ijms25052788}, 
     \cite{10560503}, \cite{Yang2023.12.03.569371} \\
\hline
HGNN & \cite{10.1093/bib/btae159}, 
     \cite{10.1093/bib/bbad391} \\
\hline
GIN & \cite{10.1093/bioinformatics/btad703} \\
\hline
General GNN & \cite{10822233},
             \cite{10.3389/fbinf.2023.1164482} \\
\hline
Hybrid & \cite{Alharbi2024}, 
         \cite{Zhou2024} \\
\hline
\end{tabular}
\label{tab:classification_subtyping}
\end{table}

\paragraph{GCN}
Graph Convolutional Networks (GCNs) have emerged as potent tools for cancer classification and subtyping due to their capacity to model the complex, non-Euclidean relationships that arise in heterogeneous biological data. Researchers are increasingly leveraging GCNs to integrate diverse multi-omics data, construct biologically meaningful graph representations, and ultimately enhance predictive accuracy across a wide spectrum of cancer types (see Table~\ref{tab:classification_subtyping_gcn}).

Multi-view GCN frameworks, such as those proposed by Ren et al. \cite{Ren2024} and MRGCN \cite{10.1093/bioinformatics/btad353}, effectively integrate gene expression, methylation, and CNV data by constructing separate graphs for each omic view. These approaches address missing modalities through the joint reconstruction of both graph structures and node features. Similarly, DeepMoIC \cite{10.1186/s12864-024-11112-5} employs residual GCNs on similarity networks derived from omics autoencoders, which promotes robust feature propagation and enhances the model's ability to capture complex biological relationships. In a broader pan-cancer context, ChebNet-based GCNs \cite{10.1145/3459930.3469542} successfully combine gene interaction networks with multi-omics data to extract both localized and global features, thereby improving predictive performance and biological interpretability.

Incorporating domain-specific biological knowledge further strengthens the learning process. For instance, MPK-GNN \cite{10148642} integrates multiple biological knowledge graphs using a multilayer GCN combined with contrastive learning, which enhances the representation of complex biological interactions. Similarly, MoCaGCN \cite{10803430} improves interpretability by learning causal graphs derived from omics similarity and gene interactions. In another approach, M-GCN \cite{10.3389/fgene.2022.884028} leverages SNV, CNV, and gene expression data along with HSIC-Lasso-based feature selection to achieve robust cancer subtype separation, while PNETwa \cite{HAN2025} fuses GCNs with MLPs on whole-exome and whole-genome sequencing data to extract population-specific signature genes.

To further improve generalization and representation learning, contrastive methods have been integrated into Graph Convolutional Network (GCN) frameworks. For instance, MOGCL \cite{Rajadhyaksha2023}, MoSCHG \cite{10385631}, and MKI-GCN \cite{10385389} leverage contrastive pre-training and cross-view alignment to enhance performance, particularly in label-scarce scenarios. Complementing these approaches, models such as SUPREME \cite{10.1093/nargab/lqad063} and MOGDx \cite{10.1093/bioinformatics/btae523} employ layered GCN architectures to jointly learn from heterogeneous omics data while preserving intrinsic inter-modality relationships. Together, these methods contribute to more robust and generalizable feature representations in multi-omics analyses

Beyond integrative classification, GCNs have also been applied in semi-supervised and multitask settings. The GCN-SC framework \cite{9669359,10.1093/bib/bbad081} facilitates label transfer and harmonization in single-cell multi-omics data, effectively bridging annotated and unannotated datasets. In another approach, the model proposed in \cite{10782177} jointly predicts cancer prognosis and subtype through a shared GCN that captures common gene–gene interactions, thereby integrating molecular insights with clinical outcomes. Additionally, the framework introduced in \cite{10.1186/s12859-023-05622-4} employs a patient–drug bipartite GCN to simultaneously model biological and pharmacological relationships, further enriching subtype predictions and advancing precision medicine.

Collectively, these studies underscore the versatility of GCN-based architectures in harnessing both the structural and functional dimensions of multi-omics data. This integrated approach yields improved accuracy, robustness, and interpretability in cancer subtyping, paving the way for more refined and clinically actionable insights.

\begin{table}[H]
\centering
\begin{tabular}{|l|c|c|c|c|}
\hline
\textbf{Paper} & \textbf{Genomic} & \textbf{Epigenomic} & \textbf{Proteomic} & \textbf{Transcriptomic} \\ \hline
\cite{Ren2024} & \checkmark & \checkmark &  & \checkmark \\ \hline
\cite{10.1145/3459930.3469542} & \checkmark &  &  & \checkmark \\ \hline
\cite{10.1093/bioinformatics/btad353} &  & \checkmark &  & \checkmark \\ \hline
\cite{10148642} & \checkmark & \checkmark &  & \checkmark \\ \hline
\cite{10.1186/s12859-023-05622-4} &  &  &  & \checkmark \\ \hline
\cite{10.1186/s12864-024-11112-5} & \checkmark & \checkmark & \checkmark & \checkmark \\ \hline
\cite{10.1093/bioinformatics/btae523} &  & \checkmark &  & \checkmark \\ \hline
\cite{10782177} &  &  &  & \checkmark \\ \hline
\cite{10803430} &  &  &  & \checkmark \\ \hline
\cite{Rajadhyaksha2023} &  & \checkmark &  & \checkmark \\ \hline
\cite{10.1093/nargab/lqad063} &  & \checkmark &  & \checkmark \\ \hline
\cite{9669359} &  & \checkmark &  & \checkmark \\ \hline
\cite{10.1093/bib/bbad081} &  & \checkmark &  & \checkmark \\ \hline
\cite{10385631} &  & \checkmark &  & \checkmark \\ \hline
\cite{10385389} &  & \checkmark &  & \checkmark \\ \hline
\cite{10.3389/fgene.2022.884028} &  & \checkmark &  & \checkmark \\ \hline
\cite{HAN2025} &  & \checkmark &  & \checkmark \\ \hline
\end{tabular}
\caption{Omics types used for classification and subtyping based on GCN.\label{tab:classification_subtyping_gcn}}
\end{table}

\paragraph{GAT}
Graph Attention Networks (GATs) have demonstrated considerable potential in cancer classification and subtyping by adaptively weighting neighboring nodes, which leads to more informative representation learning from biological graphs (see Table~\ref{tab:classification_subtyping_gat}). LASSO-MOGAT \cite{Alharbi2024LASSO} leverages this mechanism over protein–protein interaction (PPI) networks to enhance interpretability while integrating mRNA, miRNA, and methylation data. Similarly, MOGAT \cite{10.3390/ijms25052788} employs a multi-head GAT architecture that assigns distinct attention weights to each omics-derived graph, effectively capturing modality-specific patterns. Building on these innovations, MOSGAT \cite{10560503} introduces specificity-aware GATs with cross-modal attention to dynamically select relevant omics features at the sample level, thus improving adaptability across heterogeneous datasets. Further advancing this direction, DGP-AMIO \cite{Yang2023.12.03.569371} incorporates a triGAT module with directional edges from gene regulation networks to more accurately model upstream and downstream gene interactions, yielding superior performance in multi-disease subtyping.

\begin{table}[H]
\centering
\begin{tabular}{|l|c|c|c|c|}
\hline
\textbf{Paper} & \textbf{Genomic} & \textbf{Epigenomic} & \textbf{Proteomic} & \textbf{Transcriptomic} \\ \hline
\cite{Alharbi2024LASSO}           &                      & \checkmark         &                         & \checkmark              \\ \hline
\cite{10.3390/ijms25052788}       & \checkmark           & \checkmark         &                         & \checkmark              \\ \hline
\cite{10560503}                   &                      & \checkmark         &                         & \checkmark              \\ \hline
\cite{Yang2023.12.03.569371}      &                      & \checkmark         &                         & \checkmark              \\ \hline
\end{tabular}
\caption{Omics types used for classification and subtyping based on GAT.\label{tab:classification_subtyping_gat}}
\end{table}

\paragraph{HGNN}
Hypergraph Neural Networks (HGNNs) extend conventional GNNs by modeling high-order relationships among multiple entities, making them ideally suited to capture the complex dependencies inherent in multi-omics data (see Table~\ref{tab:classification_subtyping_hgnn}). For example, HyperTMO \cite{10.1093/bib/btae159} employs hypergraph convolution to represent intricate interactions among patient samples and integrates omics-specific hypergraphs through evidence theory, thereby enhancing classification under uncertainty. Similarly, scMHNN \cite{10.1093/bib/bbad391} leverages self-supervised contrastive learning alongside hyperedges to effectively align single-cell multi-omics representations, leading to improved cell-type classification by capturing non-pairwise dependencies across omics layers.

\begin{table}[H]
\centering
\begin{tabular}{|l|c|c|c|c|}
\hline
\textbf{Paper} & \textbf{Genomic} & \textbf{Epigenomic} & \textbf{Proteomic} & \textbf{Transcriptomic} \\
\hline
\cite{10.1093/bib/btae159} & & \checkmark & & \checkmark \\
\hline
\cite{10.1093/bib/bbad391} & & \checkmark & & \checkmark \\
\hline
\end{tabular}
\caption{Omics types used for classification and subtyping based on HGNN.\label{tab:classification_subtyping_hgnn}}
\end{table}

\paragraph{GIN}
Graph Isomorphism Networks (GINs) have demonstrated strong performance in biological classification tasks due to their exceptional ability to distinguish intricate graph structures (see Table~\ref{tab:classification_subtyping_gin}). For example, Ensemble-GNN \cite{10.1093/bioinformatics/btad703} integrates multiple GIN models within a federated learning framework to perform a binary classification task on patients. This model leverages human protein–protein interaction (PPI) networks to capture patient-specific molecular characteristics, successfully distinguishing patients with the luminalA subtype of breast cancer from other patients while preserving data privacy.

\begin{table}[H]
\centering
\begin{tabular}{|l|c|c|c|c|}
\hline
\textbf{Paper} & \textbf{Genomic} & \textbf{Epigenomic} & \textbf{Proteomic} & \textbf{Transcriptomic} \\ \hline
\cite{10.1093/bioinformatics/btad703} &  & \checkmark &  & \checkmark \\ \hline
\end{tabular}
\caption{Omics types used for classification and subtyping based on GIN.\label{tab:classification_subtyping_gin}}
\end{table}

\paragraph{General GNN}
General GNN frameworks have been adapted to model personalized and task-specific graph structures in multi-omics settings (see Table~\ref{tab:classification_subtyping_ggnn}). For instance, DPMON \cite{10822233} integrates general GNN embeddings with traditional machine learning layers into an optimized pipeline, demonstrating improved classification performance in COPD diagnosis. These models underscore the flexibility of GNNs in capturing subject-level variation and effectively enhancing classification outcomes within heterogeneous biomedical datasets.

\begin{table}[H]
\centering
\begin{tabular}{|l|c|c|c|c|}
\hline
\textbf{Paper} & \textbf{Genomic} & \textbf{Epigenomic} & \textbf{Proteomic} & \textbf{Transcriptomic} \\ \hline
\cite{10822233} & \checkmark &  &  & \checkmark \\ \hline
\cite{10.3389/fbinf.2023.1164482} &  & \checkmark &  & \checkmark \\ \hline
\end{tabular}
\caption{Omics types used for classification and subtyping based on general GNN.\label{tab:classification_subtyping_ggnn}}
\end{table}

\paragraph{Hybrid}
The authors of \cite{Alharbi2024} evaluate multiple graph neural network (GNN) architectures—including GCN, GAT, and Graph Transformer—for pan-cancer subtype classification. By integrating multi-omics data via LASSO, the study systematically compares performance across diverse graph structures, aiming to identify the most effective approach for capturing cancer heterogeneity.
In \cite{Zhou2024}, the MSDST method integrates GCNs with a Graph Transformer module to construct learned graphs based on omics similarities. The resulting embeddings are clustered to identify digestive tumor subtypes with distinct clinical profiles (see Table~\ref{tab:classification_subtyping_hybrid}). Together, these studies underscore the potential of hybrid GNN frameworks in leveraging multi-omics data to derive clinically actionable insights in cancer subtyping.

\begin{table}[H]
\centering
\begin{tabular}{|l|c|c|c|c|}
\hline
\textbf{Paper} & \textbf{Genomic} & \textbf{Epigenomic} & \textbf{Proteomic} & \textbf{Transcriptomic} \\ \hline
\cite{Alharbi2024}         &                      & \checkmark         &                         & \checkmark              \\ \hline
\cite{Zhou2024}            &                      & \checkmark         &                         & \checkmark              \\ \hline
\end{tabular}
\caption{Omics types used for classification and subtyping based on hybrid approaches.\label{tab:classification_subtyping_hybrid}}
\end{table}

\subsubsection{Survival analysis}
Survival analysis predicts patient risk over time by estimating hazard functions or by stratifying patients into discrete risk groups. Integrating multi-omics data via Graph Neural Networks (GNNs) enables these models to incorporate both biological interactions and patient similarity structures. This comprehensive approach captures complex interdependencies among molecular features, thereby enhancing the accuracy and interpretability of survival predictions (see Table~\ref{tab:gnn_omics_survival}).

\begin{table}[H]
\centering
\caption{GNN types and omics data used for survival analysis.}
\begin{tabular}{|c|l|c|c|c|c|}
\hline
\textbf{Paper} & \textbf{GNN Type} & \textbf{Genomic} & \textbf{Epigenomic} & \textbf{Proteomic} & \textbf{Transcriptomic} \\
\hline
\cite{ZHANG20231} & GCN & \checkmark & \checkmark &  & \checkmark \\ \hline
\cite{9669797} & GCN &  & \checkmark &  & \checkmark \\ \hline
\cite{10.3390/diagnostics14192178} & GAT &  & \checkmark &  & \checkmark \\ \hline
\cite{Yan2024} & Multi-level GNN & \checkmark & \checkmark &  & \checkmark \\ \hline
\cite{10.1093/bioinformatics/btac088} & Multi-level GNN &  &  & \checkmark & \checkmark \\ \hline
\cite{Zhu2023} & GGNN & \checkmark & \checkmark &  & \checkmark \\ 
\hline
\end{tabular}
\label{tab:gnn_omics_survival}
\end{table}

\paragraph{GCN}
GCN-based models have demonstrated strong potential in survival analysis by effectively capturing complex gene interactions and pathway dependencies. For example, LAGProg \cite{ZHANG20231} enhances local neighborhood information in sparse gene networks using CVAE-generated features, which results in improved generalization for breast cancer survival prediction. Similarly, GraphSurv \cite{9669797} integrates GCN-derived embeddings with a deep Cox model, leveraging KEGG pathways and gene–gene interactions to predict survival risk across multiple cancer types. Collectively, these approaches highlight the effectiveness of GCNs in modeling structured omics data, leading to robust and clinically meaningful survival outcome predictions.

\paragraph{GAT}
Graph Attention Networks (GATs) have demonstrated significant potential for survival analysis by selectively targeting the most critical features and interactions within omics data. In \cite{10.3390/diagnostics14192178}, a GAT-based model is utilized to predict survival outcomes in non-small cell lung cancer. By employing an attention mechanism, the model prioritizes omics features and interactions that are most relevant to patient prognosis, thereby achieving high predictive accuracy alongside enhanced biological interpretability.

\paragraph{Multi-level GNN}
Multi-level Graph Neural Networks (GNNs) have emerged as powerful tools in survival analysis by aggregating hierarchical omics features, thereby enhancing prediction accuracy and biological relevance. For instance, Yan et al. \cite{Yan2024} introduce a Multi-level GNN framework that aggregates gene-level features into pathway-level embeddings for survival risk prediction. This approach utilizes PCA initialization and incorporates pathway knowledge, which bolsters both learning efficiency and interpretability. Similarly, MLA-GNN, presented in \cite{10.1093/bioinformatics/btac088}, employs a multi-level attention mechanism to process co-expression gene modules through hierarchical GNN layers. This method not only supports glioma grading but also improves survival analysis by emphasizing gene modules that are most relevant to disease progression.

\paragraph{GGNN}
Geometric Graph Neural Networks (GNNs) capitalize on the intrinsic geometric properties of graph structures to enhance survival analysis. In \cite{Zhu2023}, a geometric GNN framework is introduced that integrates omics data with graph curvature features. By leveraging the geometric characteristics of protein–protein interaction (PPI) networks, the model achieves significant improvements in survival prediction, demonstrating state-of-the-art performance across ten cancer types.

\subsubsection{Biomarker discovery}
Biomarker discovery aims to identify molecular signatures—such as genes, proteins, or pathways—that serve as key indicators for disease diagnosis, prognosis, or therapeutic response. Graph Neural Networks (GNNs) offer a powerful framework for capturing context-aware gene relationships in multi-omics data. By modeling the complex interplay between molecular entities, GNNs facilitate robust biomarker identification, thereby advancing precision medicine (see Table~\ref{tab:gnn_omics_biomarker}).

\begin{table}[H]
\centering
\caption{GNN types and omics data used for biomarker discovery.}
\begin{tabular}{|c|l|c|c|c|c|}
\hline
\textbf{Paper} & \textbf{GNN Type} & \textbf{Genomic} & \textbf{Epigenomic} & \textbf{Proteomic} & \textbf{Transcriptomic} \\
\hline
\cite{Tan2025AMOGEL} & Hybrid &  & \checkmark &  & \checkmark \\ \hline
\cite{10.3389/fgene.2024.1513938} & GCN &  & \checkmark &  & \checkmark \\ \hline
\cite{SU2023107252} & GCN &  & \checkmark &  & \checkmark \\ \hline
\cite{10.1093/bib/bbaf108} & GraphSAGE & \checkmark &  & \checkmark &  \\ \hline
\cite{10.1093/bib/bbae658} & HGNN & \checkmark & \checkmark &  & \checkmark \\ \hline
\cite{Zhang2024.08.01.606222} & General GNN &  & \checkmark &  & \checkmark \\ \hline
\cite{10.1186/s12859-024-06015-x} & General GNN &  & \checkmark &  & \checkmark \\ 
\hline
\end{tabular}
\label{tab:gnn_omics_biomarker}
\end{table}

\paragraph{GCN}
Graph Convolutional Networks (GCNs) have proven effective for biomarker discovery by leveraging the intricate molecular interactions and hierarchical graph structures inherent in multi-omics data. In \cite{10.3389/fgene.2024.1513938}, a Multi-Level Attention GNN (MLA-GNN) is employed for liver cancer, combining gene expression and DNA methylation data to uncover pathway-level biomarkers. The innovative aspects of this model lie in its use of Cartesian product integration and hierarchical graph learning, which together enable the identification of critical gene modules that may serve as robust biomarkers.
In contrast, \cite{SU2023107252} proposes DA-SRN, an approach that optimizes sample similarity networks specifically for classification and biomarker selection. By incorporating a GCN alongside a genetic algorithm for feature refinement, DA-SRN emphasizes the optimization of the graph structure itself. This not only enhances biomarker interpretability but also boosts classification accuracy by refining the features used in the network.

Together, these studies underscore the versatility of GCN-based frameworks in biomarker discovery, demonstrating that sophisticated graph-based techniques—whether through hierarchical learning or graph structure optimization—can provide valuable insights into disease mechanisms and improve clinical decision-making.

\paragraph{GraphSAGE}
GraphSAGE, a graph neural network model designed for inductive learning, has demonstrated considerable promise in oncology for biomarker discovery by effectively capturing complex relationships within patient data. By learning to aggregate neighborhood information, GraphSAGE generalizes beyond the training set, making it particularly adept at identifying biomarkers that are critical for cancer prognosis. For instance, in \cite{10.1093/bib/bbaf108}, the authors leverage GraphSAGE to construct a patient similarity graph from heterogeneous clinical and genomic data, where nodes represent individual patients and edges capture similarities based on clinical attributes. Integrating this graph structure into a Cox proportional hazards model enhances predictive performance, as the multilayer architecture of GraphSAGE facilitates the aggregation of features from neighboring nodes to pinpoint important genes associated with patient outcomes. Metrics such as the mean hazard ratio (MHZ) and its reciprocal (RMHZ) are employed to discover genes whose expression levels significantly correlate with survival. Overall, this approach highlights how graph-based models, and GraphSAGE in particular, can be instrumental in advancing biomarker discovery and improving cancer prognosis prediction.

\paragraph{HGNN}
Hypergraph neural networks (HGNNs) have emerged as a promising approach for biomarker discovery by modeling higher-order relationships and capturing complex interactions that go beyond traditional pairwise connections. In \cite{10.1093/bib/bbae658}, the authors introduce MORE, a hypergraph-based integration network that constructs hyperedges to represent multi-omics correlations and intricate gene–gene interactions. By employing a hypergraph convolutional network, MORE is able to identify enriched modules and robust biomarkers—demonstrating its capacity to capture functionally coherent interactions that conventional pairwise graph models may overlook. This novel approach significantly enhances biomarker discovery by effectively representing the sophisticated, higher-order relationships inherent in biological data.

\paragraph{General GNN}
The authors of \cite{Zhang2024.08.01.606222} propose mosGraphGPT, a generative AI framework that constructs multi-omic signaling graphs and identifies disease‐relevant targets using attention‐based GNN layers. By focusing on critical signaling nodes and pathways, the framework facilitates biomarker discovery through the innovative combination of generative modeling with graph‐based interpretation.

Similarly, \cite{10.1186/s12859-024-06015-x} presents MPGNN-HiLander, which employs message passing GNNs on proteomic interaction data to enable hierarchical biomarker discovery in breast cancer. This approach constructs protein communities under mutation and survival pressures, subsequently using statistical enrichment to pinpoint clinically relevant biomarkers. The integration of hierarchical graph clustering with GNN-based importance scoring distinguishes MPGNN-HiLander as a compelling strategy for identifying biomarkers that elucidate disease mechanisms.

\subsubsection{Hybrid}
The authors of \cite{Zhang2024.08.01.606222} propose mosGraphGPT, a generative AI framework that constructs multi-omic signaling graphs and identifies disease‐relevant targets using attention‐based GNN layers. By focusing on critical signaling nodes and pathways, the framework facilitates biomarker discovery through the innovative combination of generative modeling with graph‐based interpretation.

Similarly, \cite{10.1186/s12859-024-06015-x} presents MPGNN-HiLander, which employs message passing GNNs on proteomic interaction data to enable hierarchical biomarker discovery in breast cancer. This approach constructs protein communities under mutation and survival pressures, subsequently using statistical enrichment to pinpoint clinically relevant biomarkers. The integration of hierarchical graph clustering with GNN-based importance scoring distinguishes MPGNN-HiLander as a compelling strategy for identifying biomarkers that elucidate disease mechanisms.

\subsubsection{Driver gene identification}
Identifying cancer driver genes is a critical challenge in oncology, as it involves detecting genes whose alterations directly fuel tumor initiation and progression. Multi-omics Graph Neural Network (GNN) models are particularly well-suited for this task due to their ability to capture intricate functional dependencies across diverse biological layers (see Table~\ref{tab:driver_gene_combined}).

\begin{table}[H]
\centering
\caption{GNN types and omics data used for driver gene identification.}
\begin{tabular}{|l|l|c|c|c|c|}
\hline
\textbf{Paper} & \textbf{GNN Type} & \textbf{Genomic} & \textbf{Epigenomic} & \textbf{Proteomic} & \textbf{Transcriptomic} \\
\hline
\cite{Peng2024} & GCN &  & \checkmark &  & \checkmark \\ \hline
\cite{10.1038/s42256-021-00325-y} & GCN & \checkmark & \checkmark &  & \checkmark \\ \hline
\cite{10.1093/bib/bbae691} & GCN &  &  &  & \checkmark \\ \hline
\cite{Dai2024.07.21.604474} & GCN & \checkmark & \checkmark &  & \checkmark \\ \hline
\cite{10385796} & GCN &  & \checkmark &  & \checkmark \\ \hline
\cite{10.1093/gigascience/giae108} & GCN &  & \checkmark &  & \checkmark \\ \hline
\cite{10.1109/JIOT.2025.3526643} & GCN &  & \checkmark &  & \checkmark \\ \hline
\cite{10.1093/bioinformatics/btac622} & GAT &  & \checkmark &  & \checkmark \\ \hline
\cite{Yuan2022.09.16.508281} & GAT &  & \checkmark &  & \checkmark \\ \hline
\cite{10385593} & Graph Transformer &  & \checkmark &  & \checkmark \\ \hline
\cite{10.1093/bioinformatics/btac478} & GIN &  & \checkmark &  & \checkmark \\ \hline
\cite{Chatzianastasis2023} & Multi-level GNN & \checkmark & \checkmark &  & \checkmark \\ \hline
\cite{10822582} & General GNN & \checkmark &  & \checkmark &  \\ \hline
\cite{10831427} & Hybrid &  &  &  & \checkmark \\
\hline
\end{tabular}
\label{tab:driver_gene_combined}
\end{table}

\paragraph{GCN}
Graph Convolutional Networks (GCNs) have become an increasingly popular approach for identifying cancer driver genes by integrating diverse biological datasets into cohesive graph-based models. In \cite{Peng2024}, the MNGCL framework leverages multi-omics data along with various biological networks within a contrastive learning paradigm. It employs Chebyshev-GCNs to generate robust embeddings from these networks, thereby enhancing the accuracy of driver gene classification. Similarly, EMOGI \cite{10.1038/s42256-021-00325-y} maps omics features onto a protein–protein interaction (PPI) graph and utilizes layer-wise relevance propagation for interpretability, which helps elucidate the contribution of individual genes to cancer development.

DriverOmicsNet \cite{Dai2024.07.21.604474} adopts a distinct strategy by integrating diverse biological interactions—such as co-expression and regulatory relationships—into a heterogeneous graph. This graph is subsequently processed through GCN layers augmented with attention mechanisms that dynamically weigh the contribution of each interaction type. In contrast, SGCD \cite{10.1093/bib/bbae691} streamlines the conventional GCN framework by eliminating the message passing mechanism, thereby addressing the challenges posed by heterophilic networks in cancer genomics. Moreover, SGCD introduces a representation separation strategy to enhance robustness and improve feature discriminability.

Additionally, gated Graph Convolutional Networks (GCNs) augmented with Gumbel-softmax attention \cite{10385796} selectively filter noisy edges across multiple protein–protein interaction (PPI) networks, effectively pinpointing driver genes and functionally coherent gene modules. Furthermore, scGraph2Vec \cite{10.1093/gigascience/giae108} leverages GCN-based variational autoencoders to generate high-resolution gene embeddings from single-cell RNA sequencing data, thereby enabling precise gene-level inference across heterogeneous single-cell populations.

Lastly, DMGNN \cite{10.1109/JIOT.2025.3526643} synergistically combines multi-omics features with graph convolution and explainable subgraph extraction techniques to facilitate the discovery of both universal and cancer-specific oncogenes with high interpretability. Collectively, these studies underscore the power of GCNs in integrating diverse omics datasets and biological interaction networks to accurately identify cancer driver genes.

\paragraph{GAT}
Graph Attention Networks (GATs) have demonstrated significant promise in cancer driver gene identification by leveraging attention mechanisms to highlight functionally important genes across diverse biological datasets. In \cite{10.1093/bioinformatics/btac622}, the MODIG model integrates multi-dimensional omics data and protein--protein interaction (PPI) networks, using attention to focus on functionally relevant genes—thereby outperforming conventional GCN-based alternatives in driver gene identification. Similarly, \cite{Yuan2022.09.16.508281} introduces a k-hop GAT model that expands neighborhood exploration for enhanced gene-level prediction. The key innovation of this approach lies in its interpretable multi-hop attention mechanism, which enables the identification of critical regulatory genes that may be missed by 1-hop models. Collectively, these methods underscore the effectiveness of GATs in capturing complex gene interactions and improving the precision of cancer driver gene identification.

\paragraph{Graph transformer}
Graph Transformer Networks (GTNs) have emerged as a powerful approach for cancer driver gene identification due to their ability to capture long-range dependencies and complex relationships within biological networks. In \cite{10385593}, the MPIT model employs multiple GTNs to extract features from distinct protein–protein interaction (PPI) networks. By fusing gene representations using max pooling and enforcing consistency constraints, the model aligns embeddings across networks effectively. This dual strategy—leveraging attention mechanisms alongside transformer-based aggregation—enables the identification of cancer-type-specific driver genes by capturing critical gene interactions across varied biological contexts.

\paragraph{GIN}
Recent advances in graph representation learning have established Graph Isomorphism Networks (GINs) as a robust foundation for modeling intricate biological networks, particularly for identifying cancer driver genes. In \cite{10.1093/bioinformatics/btac478}, the GNN-SubNet framework leverages GINs to detect disease-specific subnetworks, incorporating a model-wide explainer to extract biologically meaningful modules.

\paragraph{Multi-level GNN}
Multi-level Graph Neural Networks (GNNs) are particularly effective for cancer driver gene identification because they capture multi-scale interactions within intricate biological networks. In \cite{Chatzianastasis2023}, EMGNN leverages a meta-graph approach to integrate multiple protein–protein interaction (PPI) graphs, thereby extracting consistent gene embeddings for robust cancer gene prediction. This multi-view strategy allows the model to assimilate diverse interaction patterns, ultimately enhancing its ability to pinpoint pivotal cancer driver genes.

\paragraph{General GNN}
Paper \cite{10822582} introduces DMGNN, a novel framework that leverages mix-moment GNN layers to integrate statistical properties with omics features. By incorporating multi-moment message passing, DMGNN effectively captures both local and global dependencies, enhancing the identification of cancer driver genes.

\paragraph{Hybrid}
Creatively, one paper \cite{10831427} introduces MHN-GCN, a two-stage hybrid model that synergistically combines Graph Convolutional Networks (GCNs), Graph Attention Networks (GATs), and GATv2 with dynamic attention mechanisms to prioritize biologically relevant features. This innovative framework enhances the identification of cancer driver genes by effectively capturing both localized interactions and broader global dependencies across complex multi-omics data.

\subsubsection{Drug response prediction}
Predicting how cancer cells respond to specific drugs is crucial for advancing precision medicine. Graph Neural Networks (GNNs) facilitate the modeling of cell–drug interactions by capturing the inherent structures of molecular graphs alongside integrating comprehensive multi-omics context (see Table~\ref{tab:drug_response_combined}).

\begin{table}[H]
\centering
\caption{GNN types and omics data used for drug response prediction.}
\begin{tabular}{|l|l|c|c|c|c|}
\hline
\textbf{Paper} & \textbf{GNN Type} & \textbf{Genomic} & \textbf{Epigenomic} & \textbf{Proteomic} & \textbf{Transcriptomic} \\ 
\hline
\cite{Pu2022} & GCN & \checkmark &  &  &  \\ \hline
\cite{10.1093/bib/bbab513} & GCN &  & \checkmark &  & \checkmark \\ \hline
\cite{10.1186/s12859-024-05987-0} & GAT & \checkmark & \checkmark &  & \checkmark \\ \hline
\cite{LI2025109511} & GraphSAGE & \checkmark & \checkmark &  & \checkmark \\ \hline
\cite{10.1186/s12967-024-05976-0} & General GNN & \checkmark &  &  & \checkmark \\ \hline
\cite{10.1093/bib/bbab457} & General GNN &  & \checkmark &  & \checkmark \\ \hline
\cite{10.1093/bib/bbad406} & General GNN &  & \checkmark &  & \checkmark \\ \hline
\cite{LIU2023213} & Multi-view GNN &  & \checkmark &  & \checkmark \\ \hline
\cite{10.1038/s41598-024-83090-3} & Hybrid & \checkmark & \checkmark & \checkmark & \checkmark \\
\hline
\end{tabular}
\label{tab:drug_response_combined}
\end{table}

\paragraph{GCN}
Graph Convolutional Networks (GCNs) have demonstrated strong performance in predicting cancer drug response by effectively integrating multi-omics data with comprehensive biological networks. In \cite{Pu2022}, CancerOmicsNet employs GCN layers over cancer-specific networks to predict drug growth rate inhibition. This model combines diverse omics data with protein–protein interaction and kinase profiles through the use of attention mechanisms and JK-Net modules, thereby enhancing feature extraction and predictive accuracy. Similarly, \cite{10.1093/bib/bbab513} utilizes the LR-GNN framework to predict molecular associations, including drug–target interactions, by leveraging link-based GCN embeddings that capture complex interaction semantics beyond traditional node-level attributes.

\paragraph{GAT}
Graph Attention Networks (GATs) have been effectively applied in cancer drug response prediction due to their ability to capture intricate, multi-scale relationships within omics data. In \cite{10.1186/s12859-024-05987-0}, the MGATAF model leverages a multi-channel GAT architecture with adaptive fusion to learn comprehensive representations by integrating both local and global interactions across various omics channels. By employing a dual-level attention mechanism, the model enhances the robustness and accuracy of drug response predictions.

\paragraph{GraphSAGE}
GraphSAGE has gained prominence in drug response prediction owing to its inductive learning capabilities, which enable the model to generalize across unseen data by effectively aggregating information from neighboring nodes. In \cite{LI2025109511}, the proposed MDMNI-DGD framework integrates GraphSAGE with Graph Attention Networks (GAT) to enable druggable gene discovery. By combining multiple views derived from gene co-expression, protein–protein interaction (PPI), and pathway data, the model employs attention-based feature fusion to enhance its generalization across diverse datasets and improve drug response predictions.

\paragraph{General GNN}
In \cite{10.1186/s12967-024-05976-0}, a general Graph Neural Network (GNN) is employed on pathway‐based graphs to predict bladder cancer immunotherapy response. The model introduces a novel metric, \emph{responseScore}, along with pathway attention mechanisms to enhance interpretability.
The authors of \cite{10.1093/bib/bbab457} propose GraphCDR, a contrastive learning-based GNN for drug–cell line interaction prediction. By maximizing agreement across augmented graph views, the framework enhances representation quality and ultimately improves prediction performance.
Similarly, \cite{10.1093/bib/bbad406} presents SMG, which pre-trains a GNN on masked node prediction tasks using omics-enriched protein–protein interaction (PPI) graphs. This pretext learning strategy is subsequently fine-tuned for drug response prediction and gene classification, thereby achieving robust generalization.

\paragraph{Multi-view GNN}
Multi-view Graph Neural Networks (GNNs) have garnered significant attention in drug response prediction due to their ability to concurrently process multiple omics data types and capture intricate interdependencies across diverse datasets. In \cite{LIU2023213}, HMM-GDAN introduces a novel framework that integrates multi-scale Graph Convolutional Network (GCN) and Graph Isomorphism Network (GIN) layers with a duplex-attention mechanism applied across various omics views. This pioneering approach is the first to exploit multi-view attention for drug response prediction, substantially boosting accuracy by leveraging the complementary information inherent in different omics modalities.

\paragraph{Hybrid}
The authors of \cite{10.1038/s41598-024-83090-3} introduce an explainable hybrid Graph Neural Network (GNN) model that integrates drug molecular graphs with cellular gene expression profiles. By fusing drug substructure embeddings and omics-level embeddings, the model captures the intricate interplay between molecular features and cellular states, offering a holistic framework for drug–cancer interaction prediction. This approach not only enhances predictive accuracy but also provides critical interpretability by elucidating the key features driving therapeutic responses.

\subsubsection{Pathway discovery}
Elucidating active pathways or gene modules is crucial for understanding disease mechanisms. Graph Neural Networks (GNNs) provide a powerful framework for identifying functionally coherent gene communities within high-dimensional omics data (see Table~\ref{tab:pathway_module_combined}).

\begin{table}[H]
\centering
\caption{GNN types and omics data used for pathway/module discovery.}
\begin{tabular}{|l|l|c|c|c|c|}
\hline
\textbf{Paper} & \textbf{GNN Type} & \textbf{Genomic} & \textbf{Epigenomic} & \textbf{Proteomic} & \textbf{Transcriptomic} \\
\hline
\cite{10.1371/journal.pbio.3002369} & GCN &  & \checkmark &  & \checkmark \\ \hline
\cite{10629044} & GAT & \checkmark & \checkmark &  & \checkmark \\ \hline
\cite{Fanfani2021.06.29.450342} & General GNN &  & \checkmark &  & \checkmark \\
\hline
\end{tabular}
\label{tab:pathway_module_combined}
\end{table}

\paragraph{GCN}
Active pathway or gene module discovery is critical for understanding disease mechanisms. Graph Convolutional Networks (GCNs) excel at this task by capturing intricate gene–gene interactions and inferring pathway activities from multi-omics data. In \cite{10.1371/journal.pbio.3002369}, scapGNN leverages variational GCNs to learn gene–cell associations from single-cell multi-omics data, enabling the computation of pathway activity scores and the extraction of phenotype-linked modules. By integrating autoencoding with pathway-level inference, scapGNN effectively identifies biologically relevant gene modules, offering deep insights into the underlying disease processes.

\paragraph{GAT}
Graph Attention Networks (GATs) have emerged as a powerful tool for pathway module discovery by focusing on the most relevant gene interactions and incorporating attention mechanisms to enhance interpretability. In \cite{10629044}, the GREMI model adopts GAT layers on co-expression modules and integrates true-class probability mechanisms to achieve robust classification. Furthermore, the model employs Monte Carlo Tree Search (MCTS) to extract interpretable subgraphs, which are subsequently aligned with biological pathways, thereby deepening our understanding of the molecular processes underlying disease.

\paragraph{General GNN}
The authors of \cite{Fanfani2021.06.29.450342} introduce SBM-GNN, a groundbreaking framework that combines Graph Convolutional Networks (GCNs) with stochastic block modeling to identify cancer driver pathways. This innovative approach not only pinpoints key genes but also reconstructs pathway architectures by capturing both short- and long-range dependencies intrinsic to biological networks.

\section{Datasets}
\label{sec:datasets}

Integrating multi-omics data is pivotal for advancing cancer research, particularly when employing Graph Neural Networks (GNNs). Several comprehensive datasets—such as those provided by The Cancer Genome Atlas (TCGA), the International Cancer Genome Consortium (ICGC), and METABRIC—have been instrumental in driving this progress. These resources offer a rich spectrum of omics modalities, including genomics, transcriptomics, proteomics, and epigenomics, which together enable the construction of intricate biological networks that capture the heterogeneity and complexity of cancer. In this section, we review the notable datasets widely utilized in this domain.

%By leveraging these multi-dimensional data sources, GNNs can effectively model the interactions among molecular entities, leading to improved identification of cancer driver genes, refined tumor subtype classification, and more accurate predictions of drug response and patient outcomes. The integration of diverse omics data through GNNs ultimately paves the way for innovative precision oncology strategies and deeper insights into disease mechanisms.

\subsection{The Cancer Genome Atlas (TCGA)}

The Cancer Genome Atlas (TCGA) is a collaborative initiative designed to perform comprehensive analyses of cancer genomics \cite{TCGAReview}. This effort involves multiple specialized centers that work cohesively to collect, process, and analyze biospecimens from cancer patients. Initially, Tissue Source Sites (TSSs) acquire the biospecimens, which are meticulously processed and verified by the Biospecimen Core Resource (BCR). The subsequent clinical data and molecular analytes are submitted to the Data Coordinating Center (DCC) and further analyzed by Genome Characterization Centers (GCCs) and Genome Sequencing Centers (GSCs). Finally, the resulting sequencing data are deposited in the DCC and made available through public databases, thereby facilitating broad access to essential cancer datasets for the research community.

TCGA employs high-throughput technologies such as RNA sequencing (RNAseq), microRNA sequencing (miRNAseq), DNA sequencing (DNAseq), SNP-based platforms, array-based DNA methylation sequencing, and reverse-phase protein arrays (RPPA) to comprehensively analyze various facets of cancer genomes. These advanced platforms generate a diverse array of data types, including gene expression profiles, mutation signatures, copy number variations, and protein expression patterns. The generated data are organized into distinct levels: raw data, which typically require special permissions to access, and processed or summarized datasets that are publicly available, ensuring broad utility for research purposes. Overall, TCGA has established itself as a vital resource for advancing cancer research and deepening our understanding of cancer biology, as detailed in Table~\ref{tab:tcga_analyte_sample}.

%\begin{table}[ht]
%\centering
%\footnotesize
%\begin{tabular}{|p{1.8cm}|p{1.8cm}|p{2.3cm}|p{2.8cm}|p{1cm}|p{1cm}|p{1cm}|p{1cm}|}
%\hline
%\textbf{bcr\_patient\_uuid} & \textbf{bcr\_sample\_barcode} & \textbf{bcr\_analyte\_barcode} & \textbf{bcr\_analyte\_uuid} & \textbf{a260\_a280} & \textbf{type} & \textbf{conc.} & \textbf{rin} \\
%\hline
%9a50e7e4-831d-489f-87d2-979e987561cc & TCGA-05-4384-01A & TCGA-05-4384-01A-01D & 91a7db58-e13a-4055-8fc5-42076836c9b6 & 1.90 & DNA & 0.15 & [NA] \\
%\hline
%9a50e7e4-831d-489f-87d2-979e987561cc & TCGA-05-4384-01A & TCGA-05-4384-01A-01R & 07fc548a-e067-45da-bce4-8bcb9d418cc4 & 1.80 & RNA & 0.15 & 7.80 \\
%\hline
%9a50e7e4-831d-489f-87d2-979e987561cc & TCGA-05-4384-01A & TCGA-05-4384-01A-01T & 36831664-5b4c-40e3-b581-366c8a3d50d2 & [NA] & Total RNA & 0.10 & 7.00 \\
%\hline
%\end{tabular}
%\caption{Sample records from the TCGA LUAD (Lung Adenocarcinoma) biospecimen analyte file, showing analyte metadata for a single patient. Columns include patient and sample identifiers, analyte types (e.g., DNA, RNA), purity measurements (A260/A280), concentrations (in µg/µL), and RNA Integrity Number (RIN). This data helps researchers assess the quality and suitability of samples for downstream molecular analyses.\label{tab:tcga_analyte_sample}}
%\end{table}

\begin{table}[ht]
	\centering
	\footnotesize
	\begin{tabular}{|p{1.8cm}|p{1.8cm}|p{2.3cm}|p{2.8cm}|p{1cm}|p{1cm}|p{1cm}|p{1cm}|}
		\hline
		\textbf{bcr-patient-uuid} & \textbf{bcr-sample-barcode} & \textbf{bcr-analyte-barcode} & \textbf{bcr-analyte-uuid} & \textbf{a260-a280} & \textbf{type} & \textbf{conc.} & \textbf{rin} \\
		\hline
		9a50e7e4-831d-489f-87d2-979e987561cc & TCGA-05-4384-01A & TCGA-05-4384-01A-01D & 91a7db58-e13a-4055-8fc5-42076836c9b6 & 1.90 & DNA & 0.15 & [NA] \\
		\hline
		9a50e7e4-831d-489f-87d2-979e987561cc & TCGA-05-4384-01A & TCGA-05-4384-01A-01R & 07fc548a-e067-45da-bce4-8bcb9d418cc4 & 1.80 & RNA & 0.15 & 7.80 \\
		\hline
		9a50e7e4-831d-489f-87d2-979e987561cc & TCGA-05-4384-01A & TCGA-05-4384-01A-01T & 36831664-5b4c-40e3-b581-366c8a3d50d2 & [NA] & Total RNA & 0.10 & 7.00 \\
		\hline
	\end{tabular}
	\caption{Sample records from the TCGA LUAD (Lung Adenocarcinoma) biospecimen analyte file, showing analyte metadata for a single patient. Columns include patient and sample identifiers, analyte types (e.g., DNA, RNA), purity measurements (A260/A280), concentrations (in µg/µL), and RNA Integrity Number (RIN). This data helps researchers assess the quality and suitability of samples for downstream molecular analyses.\label{tab:tcga_analyte_sample}}
\end{table}

\subsection{STRING database}

The STRING database \cite{Szklarczyk2023} integrates protein–protein interactions (PPIs) and functional associations from various sources, including text mining, co-expression, and experimental data. It is designed to help researchers analyze PPIs and functional networks across sequenced genomes. STRING version 12.0 introduces several improvements, such as the ability to analyze full interaction networks for novel genomes, enhanced co-expression prediction using variational auto-encoders, and improved confidence scoring for experimentally derived interactions. The database employs a confidence score system to quantify the reliability of PPIs based on evidence types like genomic context, co-expression, experiments, and text mining. Additionally, STRING offers tools for functional enrichment analysis, now extended to user-submitted genomes. A key feature is its deep learning-based relation extraction model, which utilizes a RoBERTa-large architecture to identify interactions, including those that span sentence boundaries. This continually expanding data repository enables users to explore and analyze protein interactions across a wide range of biological contexts \ref{tab:ppi_sample}.

\begin{table}[ht]
\centering
\scriptsize
\begin{tabular}{|p{1.5cm}|p{1cm}|p{1cm}|p{1cm}|p{1cm}|p{1.5cm}|p{1.5cm}|p{1.5cm}|p{1.5cm}|}
\hline
\textbf{BioGRID ID} & \textbf{Entrez A} & \textbf{Entrez B} & \textbf{BioGRID A} & \textbf{BioGRID B} & \textbf{Symbol A} & \textbf{Symbol B} & \textbf{Organism A} & \textbf{Organism B} \\
\hline
103 & 6416 & 2318 & 112315 & 108607 & MAP2K4 & FLNC & Homo sapiens & Homo sapiens \\
\hline
117 & 84665 & 88 & 124185 & 106603 & MYPN & ACTN2 & Homo sapiens & Homo sapiens \\
\hline
183 & 90 & 2339 & 106605 & 108625 & ACVR1 & FNTA & Homo sapiens & Homo sapiens \\
\hline
\end{tabular}
\caption{Subset of human Protein-Protein Interaction (PPI) data from BioGRID, showing interaction records between gene products. Each row indicates a physical interaction between two proteins, identified by Entrez Gene IDs and BioGRID IDs, along with their official gene symbols and organism name. This structured format enables downstream graph-based modeling of molecular interaction networks.}
\label{tab:ppi_sample}
\end{table}

\subsection{GTEx}

The Genotype-Tissue Expression (GTEx) project \ref{tab:gtex-expression}, particularly its version 8 dataset, provides a comprehensive resource for understanding how genetic variants influence gene expression and splicing across human tissues \cite{GTEx-1}. By analyzing over 15,000 RNA-sequencing samples from 49 tissues in 838 individuals, the study by Aguet et al. delivers deep insights into both cis- and trans-regulatory associations, demonstrating that nearly all genes are subject to modulation by genetic variation. This work underscores the tissue-specific nature of these regulatory effects and highlights the importance of cell type composition in the interpretation of gene regulation. As a foundational reference, the GTEx dataset continues to drive advances in the study of gene regulation and complex traits \cite{Alasoo2023}.

\begin{table}[htbp]
\centering
\begin{tabular}{|l|l|r|r|}
\hline
\textbf{Sample ID} & \textbf{Replicated Sample ID} & \textbf{A1CF} & \textbf{A1BG} \\
\hline
GTEX-ZYT6-0626-SM-5E45V & GTEX-ZYT6-0626-SM-5E45V & 14.33 & 16.25 \\ \hline
GTEX-131YS-1626-SM-5HL6C & GTEX-131YS-1626-SM-5HL6C & 14.33 & 15.68 \\ \hline
GTEX-QEG4-1826-SM-4R1JN & GTEX-QEG4-1826-SM-4R1JN & 14.19 & 15.74 \\
\hline
\end{tabular}
\caption{
Excerpt of GTEx RNA-Seq data showing gene expression levels for the genes \textbf{A1CF} and \textbf{A1BG} across selected human tissue samples. The values, obtained from UCSC Xena, represent normalized expression levels (e.g., log$_2$(TPM+1)), which provide a standardized measure of gene activity. This type of data is crucial for studying tissue-specific expression patterns and for elucidating the relationship between gene activity and various phenotypic traits or disease states.
}
\label{tab:gtex-expression}
\end{table}

\subsection{KEGG}
\begin{figure}[htb]
    \centering
    \includegraphics[width=\textwidth]{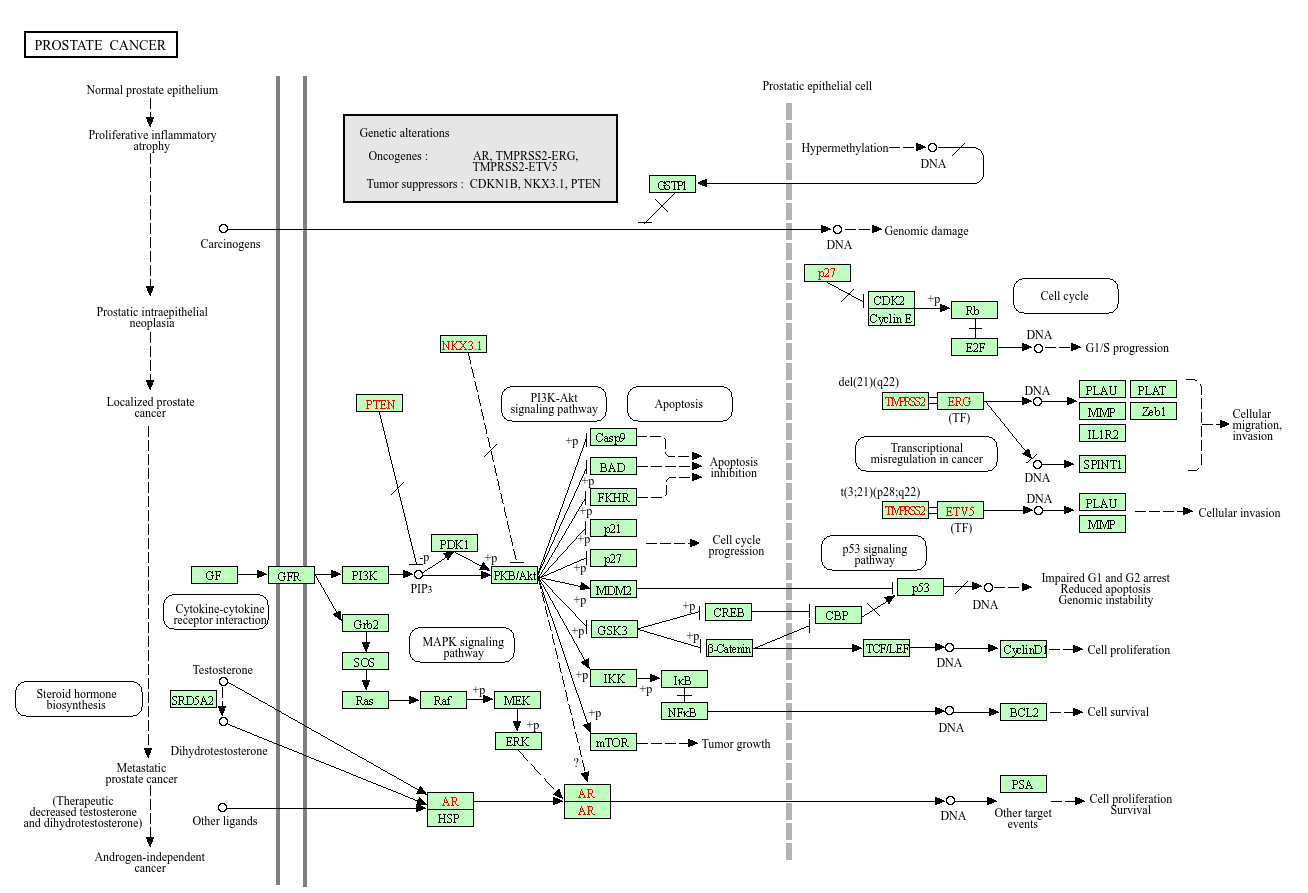}
    \caption{Homo pepiens (human) prostate cancer pathway from KEGG.}
    \label{fig:kegg}
\end{figure}

KEGG (http://www.genome.ad.jp/kegg/) is a comprehensive suite of databases and associated software tools designed for understanding and simulating higher-order functional behaviours of cells and entire organisms based on their genomic information. First, KEGG systematizes and computerizes data and knowledge on protein interaction networks through its PATHWAY database, as well as on chemical reactions via the LIGAND database, both of which are essential for deciphering various cellular processes. Second, KEGG seeks to reconstruct protein interaction networks for all organisms with completely sequenced genomes by leveraging the GENES and SSDB databases (see Figure~\ref{fig:kegg}). Third, the platform serves as a reference resource for functional genomics and proteomics experiments, notably through the EXPRESSION and BRITE databases. In this review, I will examine the current state of KEGG and highlight new developments in graph representation and graph computations that further enhance its utility in modern biological research \cite{KEGG}.

\section{Evaluation metrics}
\label{sec:metrics}

In this section, we present various evaluation metrics used to assess the performance of learning models in the context of multi-omics data analysis. These metrics can be categorized into two main groups: general metrics, which apply broadly to classification tasks, and domain-specific metrics, which are tailored to the unique characteristics of omics data and biological networks.
We begin by discussing general evaluation metrics, such as precision, recall, and F1-score, which are fundamental in measuring the accuracy and robustness of predictive models. Following that, we explore more specialized metrics, including those specifically designed for omics data, such as the PPI network distance metric and the concordance index.

\subsection{General metrics}
The general metrics discussed in this section are widely used to evaluate the performance of machine learning models across various domains, including multi-omics analysis. These metrics are considered "general" because they apply to a broad range of classification problems, independent of the specific domain or data type. They provide insights into a model's ability to correctly classify instances, handle class imbalances, and balance the trade-off between precision and recall. While these metrics are applicable to any classification task, they are not tailored to domain-specific nuances, such as biological or clinical considerations. We will explore key evaluation measures such as weighted precision, recall, F1-score, and AUPRC, each of which offers distinct advantages depending on the nature of the data and the problem at hand. These metrics are essential for understanding model performance in a comprehensive and interpretable manner.

\subsubsection{Weighted precision}
Precision measures the proportion of true positive predictions among all positive predictions made by the model:
\[
\text{Precision} = \frac{TP}{TP + FP},
\]
where \( TP \) is the number of true positives and \( FP \) is the number of false positives. High precision indicates that the model has a low false positive rate, meaning that most of the positive predictions are indeed correct.
Weighted precision, on the other hand, accounts for class imbalance by computing a support-weighted average of the precision for each class. It is defined as:
\[
\text{Weighted Precision} = \sum_{i=1}^{K} \left( \frac{n_i}{N} \cdot \text{Precision}_i \right),
\]
where:
\( K \) is the number of classes,
\( n_i \) is the number of true instances of class \( i \), and
\( N \) is the total number of instances across all classes.

This approach ensures that classes with more instances contribute proportionally more to the overall metric, which is especially important when dealing with imbalanced datasets where some classes are underrepresented \cite{SOKOLOVA2009427}.

\subsubsection{Weighted recall}
Recall, also known as sensitivity, measures the proportion of actual positives that were correctly identified:
\[
\text{Recall} = \frac{TP}{TP + FN}
\]
where \( TP \) is the number of true positives and \( FN \) is the number of false negatives. Recall is crucial when false negatives are more costly than false positives, as it reflects the model's ability to identify all relevant instances.

Weighted recall is calculated as the support-weighted average of recall scores across all classes:
\[
\text{Weighted Recall} = \sum_{i=1}^{K} \left( \frac{n_i}{N} \cdot \text{Recall}_i \right)
\]
where \( K \) is the number of classes, \( n_i \) is the number of true instances of class \( i \), and \( N \) is the total number of instances. This formulation ensures that the metric adequately reflects performance on imbalanced datasets by giving proportionately greater importance to classes with more instances \cite{SOKOLOVA2009427}.

\subsubsection{Wighted F1-Score}
The F1 score is defined as the harmonic mean of precision and recall, providing a single measure that balances both:
\[
\text{F1 Score} = 2 \cdot \frac{\text{Precision} \cdot \text{Recall}}{\text{Precision} + \text{Recall}}
\]
This metric is particularly useful when dealing with imbalanced datasets, as it penalizes models that achieve high precision at the expense of recall (or vice versa). In other words, if either precision or recall is low, the F1 score will also be low, ensuring a balanced evaluation of positive predictions. This makes the F1 score a robust performance measure in scenarios where both false positives and false negatives carry significant implications \cite{chinchor1992muc}.

%\subsubsection*{Weighted F1 Score}
Weighted F1 score calculates the F1 score for each class and averages them using the support as weights:
\[
\text{Weighted F1 Score} = \sum_{i=1}^{K} \left( \frac{n_i}{N} \cdot \text{F1}_i \right)
\]
where \( K \) represents the total number of classes, \( n_i \) is the number of true instances for class \( i \), \( N \) is the total number of instances, and \( \text{F1}_i \) is the F1 score computed for class \( i \). This formulation ensures that the evaluation reflects the underlying class distribution by giving greater importance to classes with more instances. Such a balanced evaluation is particularly beneficial when dealing with imbalanced datasets, as it prevents dominant classes from overshadowing the performance on less represented classes \cite{forman2003extensive}, as demonstrated in \cite{10.1145/3459930.3469542}.

\subsubsection{AUPRC}

The {\em Area Under the Receiver Operating Characteristic Curve (AUROC)} is a widely used metric to evaluate the performance of binary classifiers \cite{FAWCETT2006861}. It quantifies the ability of a model to distinguish between the positive and negative classes across all possible classification thresholds.

The {\em Receiver Operating Characteristic (ROC) curve} plots the {\em True Positive Rate (TPR)} against the {\em False Positive Rate (FPR)} at various threshold settings. These rates are defined as follows:
\[
\text{TPR} = \frac{\text{True Positives}}{\text{True Positives} + \text{False Negatives}}, \quad
\text{FPR} = \frac{\text{False Positives}}{\text{False Positives} + \text{True Negatives}}
\]

The AUROC is the area under this ROC curve, which can be interpreted as the probability that a randomly chosen positive instance is ranked higher than a randomly chosen negative instance by the classifier.
\begin{itemize}
	\item An AUROC of 0.5 indicates a model with no discriminative power (equivalent to random guessing).
	\item An AUROC of 1.0 indicates a perfect classifier.
	\item AUROC values closer to 1.0 indicate better classification performance.
\end{itemize}
AUROC is especially useful in imbalanced datasets because it evaluates model performance across all classification thresholds, making it inherently robust to variations in class distribution. Since it measures the probability that a randomly chosen positive instance is ranked higher than a randomly chosen negative instance, the AUROC reflects the overall discriminative ability of the classifier without being swayed by the abundance or scarcity of instances in any particular class.

The Area Under the Precision-Recall Curve (AUPRC) is a performance metric used to evaluate binary classifiers, especially in scenarios with imbalanced datasets where the positive class is rare. Unlike the Area Under the Receiver Operating Characteristic Curve (AUROC), which considers both true positives and true negatives, AUPRC focuses solely on the positive class by plotting precision (positive predictive value) against recall (sensitivity) across various threshold settings.

AUPRC is particularly informative when dealing with rare events, as it does not incorporate true negatives into its calculation. This provides a more accurate reflection of a model's ability to identify positive instances in contexts where the event of interest occurs infrequently. Consequently, AUPRC is a preferred metric in fields such as species distribution modeling and medical diagnostics. For instance, \cite{sofaer2019area} demonstrated that AUPRC offers a robust assessment of model performance for rare binary events, outperforming AUROC in scenarios with a low prevalence of the positive class.

Mathematically, AUPRC is computed as the area under the precision-recall curve, which plots precision (i.e., positive predictive value) against recall (i.e., sensitivity) at various threshold levels. This area can be interpreted as the average precision across all recall levels, providing a single scalar value that summarizes the trade-off between precision and recall for the classifier. For instance, in \cite{Chatzianastasis2023}, AUPRC was the key metric for assessing classification quality in an imbalanced dataset—an issue commonly encountered in biological problems like cancer gene prediction.

\subsection{Domain-specific metrics}
While there are many ways to evaluate the results using general techniques—such as precision, recall, and F1-score—there are certain metrics specifically designed to address the unique characteristics of omics data and the complex interactions between biological entities, such as genes, proteins, and other molecular components. These domain-specific metrics take into account the intricacies inherent in omics datasets, including the relationships within protein–protein interaction (PPI) networks, gene expression patterns, and the biological significance of various molecular features. Unlike general evaluation measures, these specialized metrics provide a more nuanced assessment of model performance in biological contexts, thereby enabling deeper insights into the underlying biological processes.
In this section, we explore several key domain-specific metrics commonly used in omics research:
\begin{itemize}
	\item {\em PPI Network Distance:} This metric evaluates the functional coherence within a set of genes by measuring the average distance or connectivity among them in a PPI network. Shorter distances typically suggest that the genes form a biologically meaningful module, which is crucial for understanding disease pathways or cellular functions.
	
	\item {\em Importance Scores:} These scores quantitatively assess the contribution of individual features (e.g., genes or proteins) to the predictions made by the model. By identifying which features hold the most weight, researchers can gain insights into potential biomarkers or key regulatory elements that underpin biological processes.
	
	\item {\em Concordance Index (C-index):} Widely used in survival analysis, the C-index measures the agreement between the predicted risks (or outcomes) and the actual observed events. A higher C-index indicates a better ability of the model to correctly rank individuals based on risk, making it indispensable for evaluating prognostic models in clinical and omics studies.
\end{itemize}
These domain-specific metrics are essential not only for assessing statistical model performance but also for ensuring that the results are biologically interpretable. They enable researchers to validate that the identified patterns, interactions, and predictive markers are consistent with known biological mechanisms and can provide insights into novel molecular pathways.

\subsubsection{PPI network distance metric}

A {\em Protein–Protein Interaction (PPI) network} is a graph structure where nodes represent proteins (or genes, assuming that each gene encodes a protein) and edges indicate interactions between them. These edges can be weighted to reflect the strength or confidence of the biological evidence supporting each interaction (e.g., based on co-expression, co-localization, experimental data, etc.).

In this context, the \textit{PPI network distance metric} is used to quantify the functional relatedness between genes. Formally, for a PPI network \( G = (V, E) \), where \( V \) denotes the set of genes/proteins and \( E \) the set of interactions, the distance between two nodes \( u, v \in V \) is defined as:
\[
\text{PPI Distance}(u, v) = \min_{\text{paths from } u \text{ to } v} \sum_{i=1}^{k} w(e_i)
\]
Here, \( e_i \in E \) represents the edges along a given path from \( u \) to \( v \), \( k \) is the number of edges in the path, and \( w(e_i) \) is the weight associated with the edge \( e_i \). Often, these weights are computed as an inverse function of the edge confidence score (e.g., \( w(e_i) = 1/\text{combined\_score}(e_i) \)), so that interactions with higher confidence yield lower weights, thereby emphasizing stronger functional connections.
This metric provides a quantitative measure of how closely related two genes are within the network, offering insights into functional modules, pathway relationships, and potentially disease-associated clusters.

In unweighted graphs, where confidence is not considered, the PPI distance simplifies to the length of the shortest path—that is, the number of edges connecting two proteins.
This metric provides a measure of biological proximity:
\begin{itemize}
	\item \textbf{Smaller distances} indicate a strong likelihood that the genes/proteins participate in the same or related pathways.
	\item \textbf{Larger distances} (or disconnected nodes) suggest a weaker or unknown functional relationship.
\end{itemize}

In recent studies, this metric has been employed to evaluate the biological coherence of gene sets selected by various methods \cite{10.1093/bioinformatics/btac478}. For example, Menche et al. \cite{PPIdistance} argue that shorter distances within the protein–protein interaction (PPI) network indicate more robust disease modules, reflecting a higher degree of functional relatedness among the genes. Similarly, in the GNN-SubNet study, the authors observed that traditional machine learning models—such as random forests and neural networks—tend to select genes that perform well individually but are often dispersed across the PPI network. This dispersion may compromise the collective biological interpretability of the selected gene sets, as the lack of close interconnectivity can weaken the evidence that these genes function together in a specific pathway or biological process.

\subsubsection{Importance score}

An {\em importance score} quantifies the contribution of a specific feature (e.g., a gene, pathway, or edge in a graph) to the prediction of a machine learning model. In GNNs, such scores are commonly derived from explainability techniques such as GNNExplainer \cite{ying2019gnnexplainergeneratingexplanationsgraph} or Integrated Gradients \cite{subramanian2020multi}. 

\paragraph{Gene-level importance}  
In GNNExplainer \cite{ying2019gnnexplainergeneratingexplanationsgraph}, a feature mask is learned over the input graph, identifying which nodes, features, and edges are most critical for the model's prediction. For gene expression data, the importance of each gene is computed for each individual sample by evaluating its contribution via the learned feature mask. To obtain a robust measure of a gene's overall significance, the sample-specific importance scores are averaged across all samples. The gene-level importance score is calculated as:
\begin{equation}
	\text{Importance}_{g} = \frac{1}{N} \sum_{i=1}^{N} s_{g}^{(i)},
\end{equation}
where \( s_{g}^{(i)} \) is the importance score of gene \( g \) in sample \( i \), and \( N \) is the total number of samples. This averaging process yields a consolidated metric that reflects the general relevance of each gene in driving the model's predictions across the dataset.

\paragraph{Pathway-level importance}
Integrated Gradients attributes the prediction to each input feature by accumulating gradients along a linear path from a baseline input to the actual input. For pathways, the importance score of a pathway is typically calculated by aggregating the importance scores of the genes it contains across all samples, often using the median:
\begin{equation}
	\text{Importance}_{p} = \text{Median}_{i=1}^{N} \left( \sum_{g \in p} s_{g}^{(i)} \right)
\end{equation}
where \( p \) denotes a pathway consisting of a set of genes \( g \), and \( s_{g}^{(i)} \) is the importance score for gene \( g \) in sample \( i \).

\paragraph{Usage and interpretation}
\begin{itemize}
	\item Higher importance scores indicate greater influence on the model's decision.
	\item Genes or pathways with high importance (e.g., \(z\)-score \(> 1.96\) or \(p\)-value \(< 0.05\)) are identified as key biomarkers or regulatory elements.
	\item These scores are often used to divide patients into high- and low-score groups for downstream survival analysis, revealing associations with clinical outcomes.
\end{itemize}

\subsubsection{Concordance index (C-index)}

The {\em Concordance Index (C-index)} is a metric used to evaluate the discriminatory power of survival models, particularly in the context of right-censored data. It measures the proportion of all pairs of subjects whose predicted survival times are correctly ordered among all pairs of subjects that can be ordered.
\[
C = \frac{1}{n_{\text{pairs}}} \sum_{i \neq j} \mathbb{I} \left( \left( \hat{T}_i > \hat{T}_j \right) \land \left( T_i > T_j \right) \right) + \mathbb{I} \left( \left( \hat{T}_i < \hat{T}_j \right) \land \left( T_i < T_j \right) \right),
\]
where:
\begin{itemize}
	\item \( \hat{T}_i \) and \( \hat{T}_j \) are the predicted survival times for subjects \( i \) and \( j \), respectively.
	\item \( T_i \) and \( T_j \) are the actual survival times for subjects \( i \) and \( j \), respectively.
	\item \( \mathbb{I} \) is the indicator function, which is 1 if the condition inside is true and 0 otherwise.
	\item \( n_{\text{pairs}} \) is the total number of comparable pairs of subjects (i.e., pairs where neither subject is censored before the other).
\end{itemize}

In multi-omics studies, the C-index is used to assess the performance of models that integrate data from various omic layers (e.g., gene expression, DNA methylation, copy number alterations) in predicting patient survival outcomes. A higher C-index indicates better model discrimination \cite{Harrell1996}. This metric accounts for the ordering of survival times in pairs of patients, making it particularly useful for right-censored data. In practice, when applying the C-index within a multi-omics framework, a model that effectively captures the complex interactions between different molecular layers will tend to achieve a higher C-index, thereby providing a more reliable ranking of patient risks and better prognostic insights.

\subsubsection{P-value}

The {\em p-value} is a fundamental concept in statistical hypothesis testing. It represents the probability of obtaining test results at least as extreme as those observed, under the assumption that the null hypothesis \( H_0 \) is true. Mathematically, it is defined as:
\[
p = P(T \geq t \mid H_0) \quad \text{or} \quad p = 2 \cdot \min \left( P(T \geq t \mid H_0),\ P(T \leq t \mid H_0) \right),
\]
where:
\begin{itemize}
	\item \( T \) is the test statistic.
	\item \( t \) is the observed value of \( T \).
	\item \( H_0 \) is the null hypothesis.
\end{itemize}
A small p-value (typically less than 0.05) indicates that the observed data is unlikely to have occurred under \( H_0 \), thereby providing evidence against the null hypothesis and favoring the alternative hypothesis.

In multi-omics research, p-values are used to assess the statistical significance of associations between omic features—such as genes and proteins—and clinical outcomes. A p-value represents the probability of obtaining test results at least as extreme as those observed, under the assumption that the null hypothesis is correct. However, when testing thousands of associations simultaneously, the chance of encountering false positives increases dramatically due to multiple testing issues.
To address this, p-values often require adjustment to control the rate of false discoveries. A common approach is the Benjamini-Hochberg procedure, which controls the False Discovery Rate (FDR). This method ranks the individual p-values, assigns each a threshold based on its rank, and identifies those p-values that are statistically significant while maintaining the FDR below a preset level (e.g., 5\%). In this way, the procedure helps ensure that the proportion of false positives among the declared significant findings remains at an acceptable level \cite{Zhou2024}.

\section{Future research directions}
\label{sec:futurework}

As the field of graph neural networks  in multi-omics cancer research continues to evolve, several promising avenues have emerged that warrant further exploration. Despite significant progress in modeling complex biological interactions and integrating heterogeneous omics data, current approaches still face limitations in scalability, interpretability, and biological fidelity. Future research must address these challenges by incorporating structured prior biological knowledge, developing more interpretable and explainable GNN architectures, and refining strategies for multi-omics data integration. Additionally, a deeper understanding of omics type co-usage patterns can guide more effective and biologically meaningful combinations of datasets. This section highlights key directions that can advance the development of robust, transparent, and biologically grounded GNN-based models for cancer studies.

\subsection{Prior knowledge graph}
In cancer research using multi-omics data, the incorporation of prior knowledge graphs has become a powerful strategy to guide graph-based learning models by embedding known biological relationships. Several studies have proposed different methodologies to construct and integrate these graphs. For example, in the framework described in \cite{Yan2024}, a guidance graph is constructed to model not only intra-omics gene regulation using mRNA gene regulatory networks (e.g., SCENIC-derived transcription factor–target relations), but also cross-omics relationships such as CNV–mRNA (positive edges) and methylation–mRNA (negative edges), enabling integration of heterogeneous omics modalities. Similarly, the study in \cite{Tan2025AMOGEL} enhances a synthetic information graph by incorporating prior knowledge from protein–protein interaction (PPI) networks obtained from the STRING database, selecting only high-confidence interactions (score \(> 500\)), as well as functional similarity graphs based on KEGG pathways and Gene Ontology (GO) annotations from DAVID. These prior graphs provide biologically meaningful structural information that supplements the learned embeddings with curated biological knowledge. Another approach \cite{10148642} proposes a shared contrastive learning framework where multiple prior graphs (such as GGI, PPI, and co-expression networks) are encoded using shared graph convolutional layers, followed by pooling and projection modules to generate aligned embeddings across omics views. This design enables better generalization and robust representation learning, especially in scenarios where labeled data are limited. The inclusion of such prior graphs—whether static regulatory structures, protein interactions, or functional pathway co-memberships—not only enhances model interpretability but also improves performance by enforcing biologically plausible constraints during message passing and embedding learning.

\subsection{Interpretability and explainability}
Explainability and interpretability methods are essential for understanding how graph neural networks leverage complex multi-omics data to make biologically meaningful predictions in cancer research.
Explainability in graph neural networks  for multi-omics cancer studies is crucial to decipher the biological mechanisms underlying cancer gene predictions and to discover novel candidate genes. A popular approach to interpret GNN models involves the use of Integrated Gradients (IG), which assigns importance scores to input features by integrating gradients of the model output along a path from a baseline to the actual input. Since GNNs take two types of inputs—node features and graph connectivity—traditional IG methods are extended by decomposing the interpretation into separate analyses of node features and edges. Node feature attribution considers contributions not only from the target gene but also from its k-hop neighbors within the graph, reflecting the message-passing process inherent in GNNs. Meanwhile, edge attribution highlights critical interactions between genes, particularly in PPI networks, by interpolating edge weights from a baseline and normalizing importance scores. This framework allows evaluation of multiple PPI networks' relative contributions, revealing that different networks provide complementary and statistically distinct information for cancer gene prediction, likely due to their unique connectivity patterns associated with tumorigenesis and progression. Gene set enrichment analysis (GSEA) on feature-ranked genes further contextualizes these findings by linking important gene nodes to known cancer pathways and hallmark gene sets, emphasizing the role of epigenetic modifications such as DNA methylation alongside genomic aberrations \cite{Chatzianastasis2023}.

To complement IG, methods like GNNExplainer identify subgraphs and subsets of node features that maximize the mutual information with model predictions, thus isolating disease-relevant subnetworks and molecular functional groups within multi-omics graphs. While traditional GNNExplainer provides instance-specific explanations, modified strategies employing induced sampling over the input space facilitate model-wide explanations by learning global node importance masks. These importance scores can then be used to weight edges in PPI networks, followed by community detection algorithms like Louvain to uncover and rank disease subnetworks. This integrated interpretability approach enables both local and global insights into how GNNs leverage multi-omics data and biological network topology to classify cancer genes, supporting the identification of key pathways and gene modules driving tumor biology \cite{10.1038/s41598-024-83090-3,10.1093/bioinformatics/btac478}.

Explainability in this context also encompasses global and local perspectives: global explanations provide an overview of the model's functioning and highlight critical biological pathways and genes that contribute across cancer types, while local explanations rationalize individual predictions by analyzing causal input-output relationships. Such dual-level interpretability enhances transparency and trust in GNN models and supports downstream applications, including drug response prediction and personalized cancer therapy. Explainable AI methods further extend to identifying important subgraphs, motifs, and causal structures within graphs, often formulated through optimization frameworks grounded in mutual information or probabilistic graphical models. These advances improve the biological relevance and robustness of predictions derived from complex multi-omics data \cite{10629044,10.3389/fbinf.2023.1164482}.

Lastly, integrating expression specificity and frequency measures with GNN-extracted subgraphs deepens the biological understanding of gene roles across cancers. This multidimensional interpretability framework captures tumor heterogeneity and diverse molecular aberrations, enabling a refined characterization of cancer-associated genes and supporting novel biomarker discovery \cite{10.1109/JIOT.2025.3526643}.

\subsection{Omics type co-usage analysis}
\begin{figure}[htbp]
    \centering
    \includegraphics[width=0.5\linewidth]{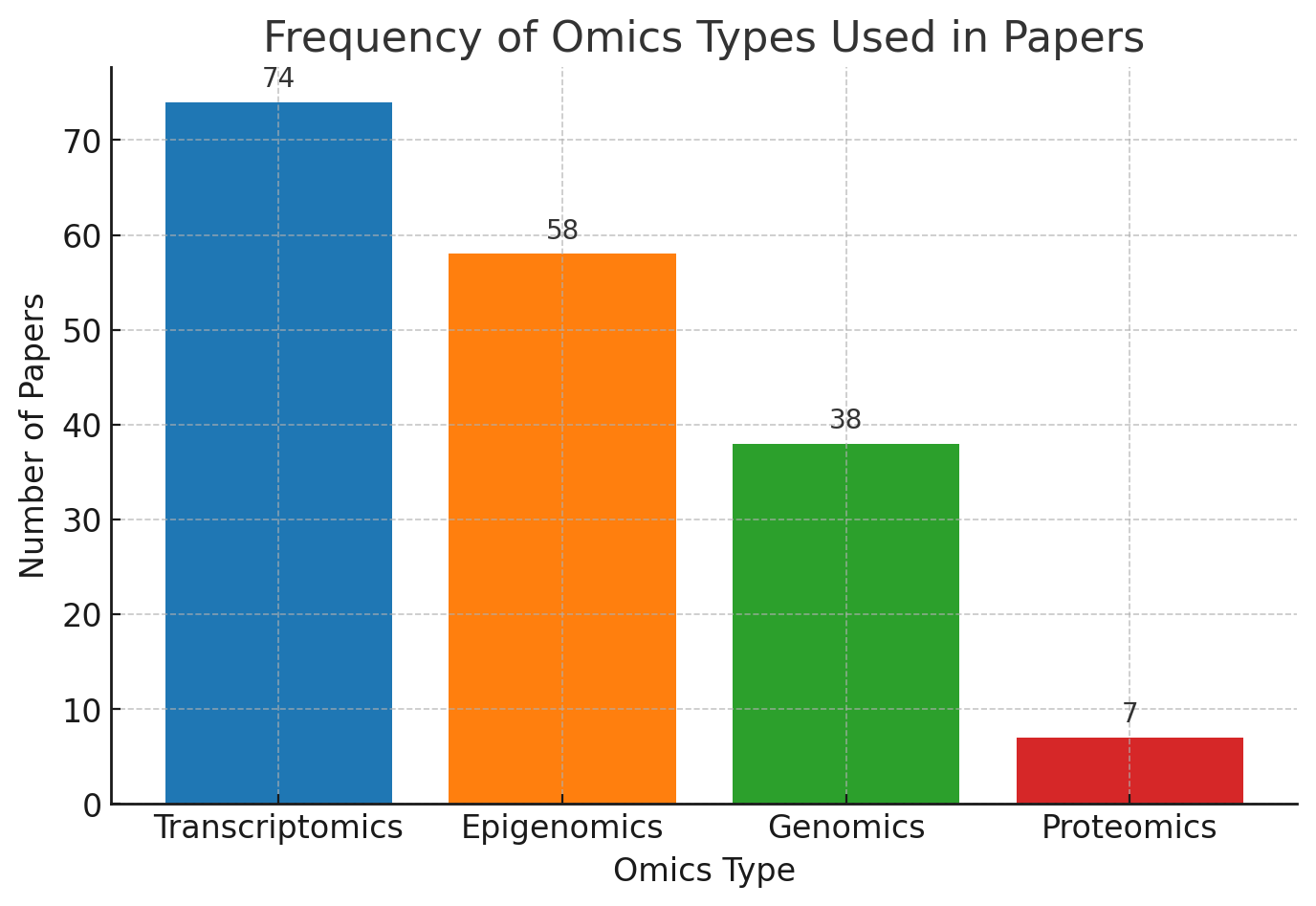}
    \caption{Number of papers using different omics types for cancer study. Transcriptomics is the most commonly used omics, followed by Epigenomics, Genomics, and Proteomics.}
    \label{fig:omics_distribution}
\end{figure}

In our comprehensive survey of 75 recent papers on multi-omics cancer studies, transcriptomics emerges as the most widely utilized data type, being incorporated in 74 of the reviewed works. Epigenomics follows closely with 58 papers leveraging this omics layer, reflecting its growing importance in understanding gene regulation and cancer progression. Genomics data is also frequently employed, appearing in 38 studies, highlighting the critical role of genetic variations in cancer biology. In contrast, proteomics is less commonly used, present in only 7 papers, possibly due to challenges related to data acquisition and integration. This distribution underscores a strong research focus on transcript-level and epigenetic alterations in cancer, while proteomic analyses remain an emerging area within multi-omics integrative approaches (Figure \ref{fig:omics_distribution}).

The analysis of multi-omics integration in recent cancer studies reveals distinct patterns in the choice and combination of data types. Transcriptomics emerges as the most widely used omics layer—featured in the vast majority of studies—followed closely by epigenomics and genomics. Notably, the high co-occurrence of transcriptomics and epigenomics, evidenced by a Jaccard similarity of approximately 0.76, indicates that these two layers are frequently combined to provide complementary insights into gene expression regulation and epigenetic modifications in cancer. Meanwhile, genomics shows a strong association with both transcriptomics and epigenomics, reflecting its critical role in characterizing underlying genetic variations in conjunction with gene activity and epigenetic states.

In contrast, proteomics displays substantially lower co-occurrence with other omics layers. This scarcity may be attributed to challenges such as large-scale protein data acquisition, variability in proteomic technologies, or the relatively nascent integration of proteomics in multi-omics frameworks. Consequently, the current research landscape predominantly focuses on integrating transcriptomic and epigenomic data, while proteomics remains an emerging dimension.

\begin{figure}[htbp]
    \centering
    \includegraphics[width=0.5\linewidth]{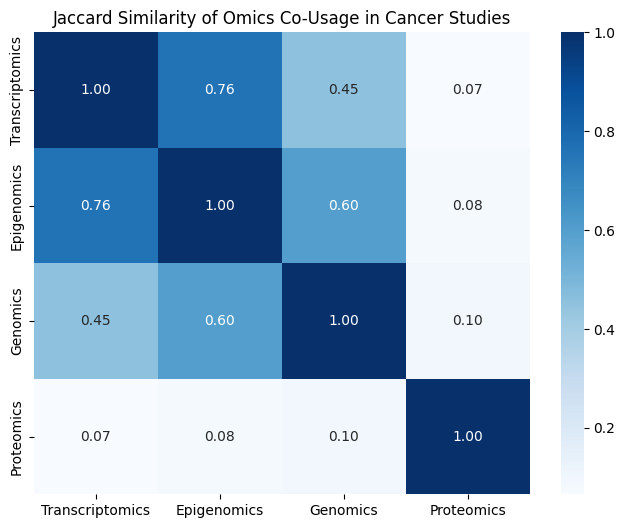}
    \caption{Jaccard similarity for co-usage of omics types.}
    \label{fig:omics_cousgae}
\end{figure}

Understanding these usage patterns not only highlights prevalent data integration strategies but also points to opportunities for future study designs. Incorporating underutilized omics types, such as proteomics, could lead to a more comprehensive molecular characterization of cancer. This broader integration could ultimately enhance our ability to decipher complex biological interactions and drive improvements in cancer diagnosis and treatment. Figure \ref{fig:omics_cousgae} provides a visual summary of these co-occurrence relationships.

\section{Methodology}
\label{sec:methodology}

To compile the literature for this survey, a structured and iterative approach was employed, primarily utilizing Google Scholar as the main database. The selection process was guided by specific inclusion criteria and a chronological framework to ensure the relevance and quality of the papers reviewed.

\subsection{Search strategy}
Initial searches were conducted using combinations of keywords such as “multi-omics”, “graph neural networks (GNNs)”, and “cancer” to identify pertinent studies. This strategy yielded an initial pool of 120 papers, which formed the basis for subsequent screening and selection processes aimed at isolating the most relevant works for our analysis.

\subsection{Chronological filtering}
The papers were then filtered based on their publication dates, starting with the most recent publications from 2025, followed by those from 2024, and then from 2021 onward. As the publication date became older, stricter selection criteria were applied to ensure the continued relevance and quality of the studies. This approach helped prioritize the most up-to-date and methodologically sound research, ensuring that our analysis reflects the current state of the field.

\subsection{Inclusion criteria}
Each paper was meticulously reviewed to ensure it met the following criteria:
\begin{itemize}
    \item Utilization of multi-omics data,
    \item Relevance to cancer research or potential applicability in cancer treatment, and
    \item Incorporation of graph neural networks in the methodology.
\end{itemize}

\subsection{Quality assessment}
Beyond the initial screening criteria, which resulted in gathering 97 papers, each of these studies was further evaluated based on additional factors such as peer reviews and citation counts. For newer studies, less emphasis was placed on citation metrics given their recent publication dates. This comprehensive review process ultimately resulted in a final selection of 75 papers that met all the established criteria for multi-omics utilization, relevance to cancer research, and integration of graph neural networks.

\subsection{Categorization}
In our survey, we adopted a strict classification approach whereby each paper was assigned to a single primary category for both task type and GNN architecture to avoid redundancy and overlapping classifications. For papers addressing multiple tasks, we identified and categorized them based on the predominant or major task addressed.
Specifically, task categorization was divided into two broad groups:
\begin{itemize}
	\item {\em Upstream Tasks:} These mainly encompass data integration and representation learning, focusing on the synthesis of heterogeneous multi-omics data into informative latent representations.
	\item {\em Downstream Tasks:} These focus on specific applications or usage scenarios, such as cancer subtype classification, survival analysis, or other clinically relevant outcome predictions.
\end{itemize}
Regarding the classification of GNN types, we exclusively considered graph neural network models and did not include non-GNN models such as multilayer perceptrons (MLPs) or other conventional machine learning architectures. When a paper employed multiple GNN variants, we assigned it to a dedicated “hybrid” class to reflect the use of combined GNN approaches.

For omics data categorization, we restricted our analysis to the four primary omics types most commonly used across the reviewed literature:
\begin{itemize}
	\item Transcriptomics,
	\item Genomics,
	\item Epigenomics, and
	\item Proteomics.
\end{itemize}
Other omics types were excluded due to their limited or negligible representation in the selected papers.
This categorization strategy ensures clarity and consistency in summarizing the literature while providing meaningful insights into the predominant trends and methodologies in multi-omics cancer research using GNNs.

\section{Conclusion}
\label{sec:conclusion}

This review systematically examined several recent studies that applied graph neural networks to multi-omics data in cancer research. The surveyed literature demonstrated the increasing effectiveness of GNNs in tasks such as cancer subtyping, prognosis prediction, and biomarker identification. A wide variety of architectures were employed, including Graph Convolutional Networks (GCNs), Graph Attention Networks (GATs), graph transformers, and hybrid or hierarchical models. The studies incorporated diverse omics layers—most notably genomics, transcriptomics, and epigenomics—and many utilized data integration strategies grounded in biological knowledge or patient similarity networks. Attention mechanisms and interpretable frameworks were frequently used to enhance model transparency. While many methods achieved promising results, challenges remained in model scalability, interpretability, and generalizability. Overall, the reviewed works underscored the potential of GNNs as a robust computational tool for advancing integrative cancer research.

\bibliographystyle{elsarticle-num}
\bibliography{references} 

\end{document}